\documentclass[namedreferences]{kluwer}    

\newdisplay{guess}{Conjecture}

\usepackage{amstext}   
\usepackage{epsf}
\usepackage{graphicx}
\usepackage{latexsym}
\usepackage{wasysym}
\usepackage{lscape}
\usepackage{times,mathptm}

\usepackage[usenames]{color}
\def\aa{\textcolor{Black}}
\def\bb{\textcolor{Black}}

\begin{document}
\begin{article}

\begin{opening}
\title{Distributional Measures as Proxies for\\
Semantic Relatedness}
\author{Saif \surname{Mohammad}\email{smm@cs.toronto.edu}\\}
\author{Graeme \surname{Hirst}\email{gh@cs.toronto.edu}}
\institute{University of Toronto, Toronto, ON M5S 3G4, Canada}

\runningtitle{Distributional Measures as Proxies for Semantic Relatedness}
\runningauthor{Saif Mohammad and Graeme Hirst}

\begin{abstract}
The automatic ranking of word pairs as per their semantic relatedness and ability
to mimic human notions of semantic relatedness has widespread
applications. Measures that rely on raw data (distributional
measures) and those that use knowledge-rich ontologies
both exist. Although extensive studies have been 
performed to compare ontological measures with human judgment,
the distributional measures have primarily been evaluated by
indirect means. This paper is a detailed study of some of the
major distributional measures; it lists their respective
merits and limitations. New measures that overcome these
drawbacks, that are more in line with the human notions of semantic relatedness, 
are suggested. The paper concludes with an exhaustive comparison
of the distributional and ontology-based measures. Along the way,
significant research problems are identified. Work on these problems
may lead to
a better understanding of how semantic relatedness is to be measured.
\end{abstract}

\keywords{Distributional similarity/relatedness, 
          semantic similarity/relatedness, 
		  word association,
		  relative entropy, 
		  asymmetric measures,
		  compositional/non-compositional measures,
		  pseudo-fuzzy metrics
}
\abbreviations{\abbrev{PCM}{Primary Compositional Measure};
               \abbrev{CRM}{Co-occurrence Retrieval Model};
               \abbrev{KLD}{Kullback-Leibler Divergence};
               \abbrev{PMI}{Pointwise Mutual Information}
}

\end{opening}

\section{Introduction}
Humans are inherently capable of determining whether one word pair is more
semantically related than another. For example, given the word pairs 
{\em honey}--{\em bee} and {\em paper}--{\em car}, one can easily identify 
the former pair to be more semantically related than the latter. This, however, is not
true for machines. A lot of work has been done in automating the
process in the last fifteen years.  While some approaches do better
than others and have been applied to solving practical problems,
none has matched human judgment. 

Typically, automated systems assign a score of {\bf semantic relatedness} to
a given pair of words ({\bf target words}) calculated from a {\bf relatedness measure}. The absolute
score is usually irrelevant on its own.  For example, a relatedness
score of 0.7 between {\em a} and {\em b}, in a possible range of 0 to
1, does not imply that {\em a} and {\em b} are more related than the
average word pair. However, given that the semantic relatedness of
{\em c} and {\em d} is 0.6, the system can conclude that {\em a} and
{\em b} are more related than {\em c} and {\em d}. Thus even though
the absolute score given by a relatedness measure is not of much
significance, it is important that the measure give a higher score to
word pairs which humans think are more related and comparatively lower
scores to word pairs that are less related. This ability to mimic
human judgment of semantic relatedness has been used in numerous
applications such as automated spelling correction, word sense
disambiguation, thesaurus creation, information retrieval, text
summarization, and identifying discourse structure. 

Existing measures of semantic relatedness rely either on
ontologies and semantic networks or just raw text. 
\inlinecite{Budanitsky99}, \inlinecite{BudanitskyH00} and \inlinecite{PatwardhanBT03} do an
extensive survey and comparison of the various
WordNet-based measures. Measures that use just raw text,
known as the {\bf distributional measures}, have been described
individually (for example, in \inlinecite{SchutzeP97}, \inlinecite{Hindle90}, 
\inlinecite{Lin98C}, \inlinecite{PereiraTL93}, etc) but have not been
extensively compared among each other.
This paper focuses on distributional measures and analyzes
their strengths and limitations. Particular attention is paid to the different
kinds of distributional measures and their components.
New measures are suggested that overcome some of
their drawbacks. Characteristics of WordNet-based and distributional 
measures are contrasted and finally, future research directions are
suggested which
may determine a better understanding of semantic relatedness.

\section{Background}

\subsection{Co-occurrences}
Words that occur within a certain window of a target word are
called the {\bf co-occurrences} of the word. The window size may be a
few words on either side, the complete sentence, a paragraph or
the entire document. Consider the sentence below:
\begin{center}
{\tt the plane flew through a cloud}
\end{center}

\noindent If we consider the window size to be the complete sentence,
{\em flew} co-occurs with {\em the, plane, through, a}
and {\em cloud}. The set of words that co-occur with a word
constitute the context of the word. They are used in tasks such as 
information retrieval, word sense disambiguation, and semantic
relatedness.
\pagebreak
\subsection{Word Association Ratio}
Given two events $x$ and $y$ with probabilities $P(x)$ and $P(y)$,
their {\bf pointwise mutual information}~\cite{Fano61}\endnote{In their respective papers, Robert Fano as well
as Ken Church and Patrick Hanks refer to pointwise mutual information as mutual
information.}, PMI for short, or just
$I$, is defined as follows:

\begin{equation}
\label{eq:MI}
I(x,y) = \log_{2}\frac{P(x,y)}{P(x)P(y)} 
\end{equation}

\noindent $P(x,y)$ is the joint probability of $x$ and $y$. If $I(x,y)$
evaluates to be close to zero, i,e, $P(x,y) \approx P(x) \times P(y)$,
then it means that events $x$ and $y$ occur together just as often 
as is expected from their individual probabilities. 
If $I(x,y) \gg 0$, it implies that $x$ and $y$
occur together more often than would be expected from their individual
probabilities and hence have a strong correlation. 

\inlinecite{ChurchH89}\endnotemark[\value{endnote}] introduce {\bf word association ratio}, which is similar 
to pointwise mutual information. If $x$ and $y$ are words with probabilities
$P(x)$ and $P(y)$ (estimated by corpus counts), their  
association ratio is defined to be the same as in (\ref{eq:MI}),
except that $P(x,y)$ stands for the probability that $x$ appears,
within a certain window, before $y$. It should be noted that 
$P(x,y)$ is no longer symmetric ($P(x,y) \neq P(y,x)$) as $P(x,y)$ and $P(y,x)$ represent
two different events. 
If two words have a word association ratio
close to zero then they do not share an interesting relationship but
if $I(w_1, w_2) \gg 0$, then $w_2$ follows $w_1$ (within a certain window)
more often than chance and the words $w_1$ and $w_2$ are strong co-occurrences.
\aa{Theoretically, word association ratio may yield negative values
(word pair occurs less frequently than expected by random chance)
but \inlinecite{ChurchH89} show that it is hard to accurately predict
negative word association ratios with confidence. Systems which
use word association ratio may be adversely affected by this. A common
approach to counter this is to equate the negative association values
to 0 (for example, \inlinecite{Lin98C}). This usually means that the system will
ignore such words.}

A problem with PMI in general (which is inherited by word association ratio)
is that low frequency events get higher scores than expected. \inlinecite{PantelL02}
try to overcome this by multiplying the PMI value with a correction factor.
 Although, Pantel and Lin give the correction factor
for word association ratio using syntactically related co-occurring words,
a more generic form applicable for pointwise mutual information is as shown below:

\begin{equation}
\label{eq:PMICor}
I_{\text{\itshape corrected}}(x,y) = \log_{2}\frac{P(x,y)}{P(x)P(y)} \times \frac{\min (\text{\itshape freq}(x),\text{\itshape freq}(y))}
                                                     {\min (\text{\itshape freq}(x),\text{\itshape freq}(y)) + 1}
\end{equation}

\noindent \aa{The correction factor is large (close to 1) if both the events occur
a large number of times and small (close to 0) if any of the two events
occurs very few times.}

\subsection{Relatedness vs Similarity} 

A closely related concept to semantic relatedness is {\bf semantic 
similarity}.  While there is some overlap in their meanings and they
may be used interchangeably in certain contexts, it is important to be
aware of their distinction.  \inlinecite{BudanitskyH00} and \inlinecite{BudanitskyH04}
point out that semantic similarity is used when
similar entities such as {\em apples} and {\em
bananas} or {\em table} and {\em furniture} are compared. 
These entities are close to each other in an is-a hierarchy.
For example, {\em apples} and {\em bananas} are hyponyms of
{\em fruit} and {\em table} is a hyponym of {\em furniture}.
However, even dissimilar
entities may be semantically related, for example, {\em door} and {\em knob}, {\em tree} and {\em shade}, or
{\em gym} and {\em weights}. In this case the two entities are 
not similar per se, but are related by some relationship. This relationship
may be one of the classical relationships such as meronymy (is part of) as in {\em door--knob}
or a non-classical one as in {\em tree--shade} and {\em gym--weights}.
Thus two entities are semantically related if they are semantically similar (close together
in the is-a hierarchy) or share any other classical or 
non-classical relationships. As \inlinecite{BudanitskyH04} point out,
semantic similarity is a subset of semantic relatedness.


The concept of {\bf semantic distance} has traditionally been used
in the context of both semantic relatedness and semantic similarity.
In the former context, it represents the inverse of semantic relatedness,
while in the latter, it is the inverse of semantic similarity. In this paper
as well, we shall continue to use the term for both concepts with
the confidence that the context will disambiguate the intended
meaning.


\subsection{The Distributional Hypothesis}


Given a text corpus, individual words have more or less differing contexts 
around them. The context of a word is composed of words co-occurring with
it within a certain window around it.
Distributional measures use statistics acquired from a large text corpora 
to determine how similar
the contexts of two words are.
These measures are also used as proxies to measures of semantic similarity
as words
found in similar contexts tend to be semantically similar.  This is known as 
the {\bf distributional hypothesis} (\inlinecite{Firth57} and \inlinecite{Harris68})
and such measures have traditionally
been referred to as measures of {\bf distributional similarity}.

The hypothesis makes intuitive sense as \inlinecite{BudanitskyH04} point out.
If two words have many co-occurring
words then similar things are being said about both of them and 
so they are likely to be semantically similar. Conversely, if two words are 
semantically similar
then they are likely to be used in a similar fashion in text and thus
end up with many common co-occurrences. For example, the semantically 
similar {\em bug} and {\em insect} are expected to have a number of common
co-occurring words such as {\em crawl, squash, small, woods}, and so on,
in a large enough text corpus.

Like measures of distributional similarity there exist measures of
what we will call {\bf distributional relatedness} (\inlinecite{SchutzeP97} and \inlinecite{YoshidaYK03}).
These measures use raw text
and co-occurrence information to determine semantic relatedness between
two words. The distributional hypothesis mentioned earlier is generic
enough to be the basis for both distributional similarity and 
distributional relatedness. We propose more specific hypotheses that
demarcate the two.

\begin{quote}
{\bf Hypothesis of distributional similarity:} \\
{\bf Distributionally similar} words tend to be semantically similar,
where two words ($w_1$ and $w_2$, say) are said to be distributionally similar if they have 
many common co-occurring words and these co-occurring words are each
related to $w_1$ and $w_2$ by the same syntactic relation.
\end{quote}

\begin{quote}
{\bf Hypothesis of distributional relatedness:} \\
{\bf Distributionally related} words tend to be semantically related,
where two words ($w_1$ and $w_2$, say) are said to distributionally related if they have 
many common co-occurring words and this set of co-occurring words is not 
restricted to only those that are related to $w_1$ and $w_2$ by 
the same syntactic relation.
\end{quote}

The two hypotheses are based on the fact that semantically similar
words belong to the same broad part of speech (noun, verb, etc)
and are thus each syntactically related to most common co-occurring 
words by the same syntactic relation. Further, the more two 
words are semantically related,
the more common co-occurring words they have.
Consider the semantically related word pair {\em doctor--operate}.
In a large enough body of text, the two words are likely to
have the following common co-occurring words: {\em patient, scalpel,
surgery, recuperate}, and so on. All these words 
will be used by a measure of distributional relatedness and the pair
will be assigned a high score. However, a measure
of distributional similarity will not use any of these co-occurring words
(and likely no other, for that matter) as they are not related to 
the target words by the same syntactic relation. The word
{\em doctor} is almost always used as a noun while {\em operate}
is a verb. Thus {\em doctor} and {\em operate} will get a very
low score of distributional similarity. The word pair {\em doctor--nurse},
on the other hand, will get a high score of distributional relatedness and
distributed similarity.
Thus an important characteristic of any
distributional measure is whether it is a measure of distributional similarity or more
generally that of distributional relatedness. 

It should be noted that a measure of distributional similarity
will provide a high score for certain closely
related but dissimilar words belonging to the same thematic role.
For example, {\em homeless} and {\em drunk} which refer to dissimilar
concepts but share a non-classical relationship of association
({\em homeless} and {\em drunk} tend to occur together in text)
will likely get a high score as they belong to the same part of 
speech (adjective) and may have many common co-occurring words
such as {\em beggar, person, helped,} and so on, related
by the same syntactic relation.
This is a limitation of the current measures of distributional
similarity and the impact of the limitation on the 
ability of the measures to mimic semantic similarity is worth determining.

The relevant literature uses the term {\bf distance} as inverse of 
distributional similarity. In order to clearly distinguish between
semantic distance, this paper will refer to the inverse of 
distributional similarity as {\bf distributional distance}.
Like semantic distance, distributional distance will also be 
used as the inverse of distributional relatedness, and the context
should help disambiguate the intended meaning. 

\subsection{Relatedness of Words and Concepts}
Measures of semantic relatedness and similarity are applied to 
particular concepts (or particular senses of the words); for example,
one may determine the semantic relatedness of
{\em bank} in the {\em financial institution} sense and {\em 
interest} in the {\em interest rate} sense. Distributional 
measures, on the other hand, usually assign scores to word pairs irrespective
of the nature of their polysemy (how many senses they have)
or the particular senses they have been used in. 
Distributional measures will need a much more knowledge 
rich source (for example large amounts of sense-tagged corpora) than raw text 
to assign scores to word-sense pairs.

\subsection{Evaluation}

The presence of a large number of relatedness measures necessitates
a suitable evaluation to determine which methods come closest
to the human notions of relatedness and to determine how good they
each are. There exist two modes of evaluation. The first 
involves the creation of two ranked lists of certain word pairs.
One list is created using a relatedness measure while the other
is ranked by humans.
The correlation of the
two rankings is indicative of how closely the measure mimics
human judgment of relatedness. \inlinecite{RubensteinG65} were the first to
conduct quantitative experiments with human subjects
who were asked to rate 65 word pairs on a scale of 0.0 to 4.0 as per 
their relatedness. The word pairs chosen ranged from very
similar and almost synonymous to unrelated. \inlinecite{MillerC91} also 
conducted a similar study on 30 word pairs taken from the Rubenstein-Goodenough
pairs. However, lack of large amounts of data from human subject experimentation
limits the quality of this mode of evaluation.

The second and a more indirect way of evaluating measures of
semantic relatedness is by the performance
of natural language tasks that use them,
for example, automatic spelling correction, word sense
disambiguation, estimation of unseen bigram \aa{(not found in training data)}
probabilities, and so on.

\section{Distributional Measures}
\subsection{Spatial Metrics}

A popular technique to determine distributional relatedness between two words
is to map them to points in a multidimensional space such that the distance
between the two points is an indicator of distributional and thereby semantic distance
between them. 

Large co-occurrence matrices pertaining to each word, which store 
the set of words that co-occur with it within a certain
window size, are created from a text corpus. 
Consider a multidimensional space where
the number of dimensions is equal to size of vocabulary. A word
$w_1$ can be represented by a point in this space such that
the vector $\vec w_1 $ from the origin to this point has 
equal positive components in all dimensions corresponding to
words that co-occur with $w_1$. Similarly, vector $\vec w_2$
can be created for word $w_2$. This section describes three distributional distance metrics
that quantify the distance between $\vec w_1 $ and $\vec w_2$.

\subsubsection{Cosine}

The {\bf cosine} method (denoted by $\text{\itshape Cos}$) is one of the earliest distributional measures.
Given two words $w_1$ and $w_2$, the cosine measure calculates
the cosine of the angle between $\vec w_1 $ and $\vec w_2 $.
If a large number of words co-occur with both $w_1$ and $w_2$,
$\vec w_1 $ and $\vec w_2$ will have a small angle between 
them, the cosine will be large, and we get a large
relatedness value between them. \aa{The cosine measure gives 
scores in the range from $0$ (unrelated) to $1$ (maximally related). }


\begin{equation}
\text{\itshape Cos}(w_1,w_2) = \frac{\vec w_1 . \vec  w_2}{\mid \vec w_1 \mid \times \mid \vec w_2 \mid} 
\end{equation}

\noindent A limitation of the cosine method in its original form is that all
co-occurring words are treated the same, irrespective of
how often they co-occurred with $w_1$ and $w_2$. 
A popular variation \aa{(\inlinecite{YoshidaYK03}, \inlinecite{Lee99}, and \inlinecite{SchutzeP97})} that
incorporates this information is stated below:

\begin{equation}
\label{eq:Cos}
\text{\itshape Cos}(w_1,w_2) = \frac{\sum_{w \in C(w_1) \cup C(w_2)} \left( P(w|w_1) \times P(w|w_2) \right) }
{\sqrt{ \sum_{w \in C(w_1)} P(w|w_1)^2 } \times \sqrt{ \sum_{w \in C(w_2)} P(w|w_2)^2 } } 
\end{equation}

\noindent $C(x)$ is the set of words that co-occur (within a certain window)
with the word $x$ in a corpus. 
\noindent \aa{$P(x|y)$ is the probability that a particular co-occurrence is composed of
$x$ and $y$, given that word $y$ is one of the words in the co-occurrence pair.}
It can be approximated by simple corpus counts.
Once again, the formula is the cosine of the angle between the word vectors
$\vec w_1 $ and $\vec w_2$ but the word vectors incorporate the strength of association
of the co-occurring words with the target words. The component of $\vec x$ in a
dimension (corresponding to word $y$, say) is equal to the strength of association 
of $y$ with $x$. Thus the vectors corresponding to two words are closer together, 
and thereby get a high distributional relatedness score, if they share many co-occurring words
and the co-occurring words have more or less the same strength of association 
with the two target words. In the above formula conditional probability
of the co-occurring words given the target words is used as the strength
of association. 


The cosine is used, among others, by \inlinecite{SchutzeP97} and 
\inlinecite{YoshidaYK03}, who suggest methods of automatically generating thesauri from 
text corpora. \inlinecite{SchutzeP97} use the
Tipster category B corpus~\cite{Tipster} (450,000 unique terms)
and the {\em Wall Street Journal} to create a large but sparse
co-occurrence matrix of 3,000 medium-frequency words (frequency
rank between 2,000 and 5,000). Latent semantic indexing and 
single-value decomposition (see \inlinecite{SchutzeP97} for details) are used to reduce the dimensionality
of the matrix and get for each term a word vector of its 20 strongest
co-occurrences. The cosine of a word vector (say $\vec w_1$) with each of the other
word vectors is calculated and the top scores along with 
the words whose vector generated the top scores is noted. These
words form the thesaurus entries for $w_1$.


\inlinecite{YoshidaYK03} believe
that words that are closely related for one person may be
distant for another. They use 
around 40,000 HTML documents to generate personalized thesauri 
for six different people. Documents used to create the thesaurus 
for a person are retrieved from the subject's home page and a web crawler which
accesses linked documents. The authors also suggest a root-mean-squared 
method to determine the similarity of two different thesaurus
entries for the same word.

\subsubsection{Manhattan and Euclidean Distances}

Distance between two points (words) in multidimensional space can 
be calculated using the {\bf Manhattan distance} a.k.a.\@ {\bf $L_1$ norm} 
(denoted by $L_1$)
or {\bf Euclidean distance} a.k.a.\@ {\bf $L_2$ norm} (denoted by $L_2$).
In the Manhattan distance~(\ref{eq:L1}) \aa{(\inlinecite{DaganLP97}, \inlinecite{DaganLP99}, 
and \inlinecite{Lee99})}, the disparity in strength of association of
$w_1$ and $w_2$ with each word that they co-occur with, is summed.
The more the disparity in association, the more is the distributional distance between the two words.
The Euclidean distance~(\ref{eq:L2}) \aa{(\inlinecite{Lee99})} employs the root mean squared of the 
disparity in association to get the final distributional distance.
\aa{Both $L_1$ norm and $L_2$ norm give values in the range 0 (zero
distance or maximally related) and infinity (maximally distant or unrelated).}

\begin{eqnarray}
\label{eq:L1}
L_1(w_1,w_2)& =& \sum_{w \in C(w_1) \cup C(w_2)} \mid P(w|w_1) - P(w|w_2) \mid \\
\label{eq:L2}
L_2(w_1,w_2)& =& \sqrt{\sum_{w \in C(w_1) \cup C(w_2)} \left(P\left(w|w_1\right) - P\left(w|w_2\right)\right)^2 }
\end{eqnarray}

\noindent The above formulae use conditional probability 
of the co-occurring words given the target words as the strength
of association.
The distributional relatedness of words may be found by taking the reciprocal of the
distributional distance or similar suitable method.


\aa{\inlinecite{Lee99} compared the ability of all three spatial metrics to
determine the probability of an unseen (not found in training data) word
pair. The measures in order of their performance (from better to worse) were: $L_1$ norm,
cosine, and $L_2$ norm.}
\aa{\inlinecite{Weeds03} determined the correlation of word pair ranking as per
a handful of distributional measures with human rankings (Miller and Charles
word pairs~\inlinecite{MillerC91}). Using verb-object pairs from the {\em British
National Corpus (BNC)},
she found the correlation of $L_1$ norm with human rankings to be 0.39.}

\subsection{Set Operations}
\aa{ Distributional measures, as discussed earlier, aim to determine semantic
similarity (or relatedness) using words that co-occur with
the target words. 
The problem can be transformed to finding the similarity of 
two sets ($W_1$ and $W_2$, say), where each set has as its members the co-occurring words of the
two target words ($w_1$ or $w_2$), respectively. 
One can now use set operations such as {\bf Jaccard} and {\bf Dice coefficient}
to determine the similarity of the two sets and thereby, the semantic 
similarity of the target words.}

\begin{eqnarray}
\label{eq:Jaccard}
\text{\itshape Jaccard}(w_1,w_2)& =& \frac{\left|W_1 \cap W_2\right|}{|W_1 \cup W_2|} \\
\label{eq:Dice}
\text{\itshape Dice}(w_1,w_2)& =& \frac{2 \times \left|W_1 \cap W_2\right|}{|W_1| + |W_2|} 
\end{eqnarray}

\noindent Both measures give
scores in the range from $0$ (unrelated) to $1$ (maximally related)
and will rank
word pairs identically.
\pagebreak
\newdisplay{theorem}{Theorem}
\begin{theorem}
\bb{If the similarity of word pair
one is less than the similarity of word pair two, as determined by
the Jaccard coefficient, then the similarity of word pair one will 
be less than the similarity of word pair two, as determined by 
the Dice Coefficient.}
\end{theorem}

\begin{pf*}{Proof}

Let $x$ be the number of co-occurrences common to 
word pair one and $y$ the number of words that co-occur with just
one of the two words in word pair one.\\
Let $l$ and $m$ be the
corresponding values for word pair two.\\
Therefore,
\begin{eqnarray}
\text{\itshape Jaccard}(\text{\itshape pair one})& =& \frac{x}{x+y} \\
\text{\itshape Dice}(\text{\itshape pair one})& =& \frac{2x}{2x+y} \\
\text{\itshape Jaccard}(\text{\itshape pair two})& =& \frac{l}{l+m} \\
\text{\itshape Dice}(\text{\itshape pair two})& =& \frac{2l}{2l+m} 
\end{eqnarray}


\noindent Given,
\begin{eqnarray}
\text{\itshape Jaccard}(\text{\itshape pair one})& <& \text{\itshape Jaccard}(\text{\itshape pair two})\\
\Rightarrow \quad \frac{x}{x+y}& <& \frac{l}{l+m}\\
\Rightarrow \quad xl+xm& <& xl+yl\\
\label{eq:condition}
\Rightarrow \quad xm& <& yl 
\end{eqnarray}
To prove,
\begin{eqnarray}
 \text{\itshape Dice}(\text{\itshape pair one})& <& \text{\itshape Dice}(\text{\itshape pair two})\\
 \text{or,}\quad \frac{2x}{2x+y}& <& \frac{2l}{2l+m}\\
 \text{or,}\quad 4xl+2xm& <& 4xl+2yl\\
 \text{or,}\quad xm& <& yl 
\end{eqnarray}
which is true (from (\ref{eq:condition})). \qed
\end{pf*}
Thus, in terms of measuring distributional similarity/relatedness,
Jaccard and Dice coefficients are identical. \inlinecite{Lee99} shows that the Jaccard
coefficient 
performs better than $L_1$ norm in an unseen bigram probability
estimation task.

\subsubsection{Pseudo-Fuzzy Metrics}
\aa{Simple set operations as stated above do not consider the strength of
association of the co-occurring word with the target words. 
The strength of association can be incorporated into the metrics
by considering the co-occurrence sets to be {\bf pseudo-fuzzy}.
The degree of membership of each word in the pseudo-fuzzy set corresponding
to a target word is its strength of association
with the target word. We call the sets pseudo-fuzzy (and not fuzzy)
because the range of membership values is now dependent on the measure of association
used --- conditional probability: 0 to 1, PMI (ignoring negative values\endnote{It 
is hard to accurately predict
negative word association ratios with confidence (\inlinecite{ChurchH89}).}): 0 to infinity. 
Even though conditional probability has a range from 0 to 1 like a 
standard fuzzy set membership function, the conditional probabilities
of all the words with respect to a particular target word sum up to $1$.
This need not be (and usually is not) true of the membership values for a regular fuzzy set.}

Use of conditional probability (denoted by CP) as the strength of association and application
of {\bf Jaccard} and {\bf Dice coefficient} on the pseudo-fuzzy set results in the following
formulae.
Similar to the case of regular sets, it can easily be shown
that the Dice and Jaccard coefficients of pseudo-fuzzy sets also rank
word pairs identically.

\begin{eqnarray}
\label{eq:JaccardCP}
\text{\itshape Jaccard}^{\text{\em CP}}(w_1,w_2)& =& \frac{\sum_{w \in C(w_1) \cup C(w_2)} \min (P(w|w_1),P(w|w_2))}
                                       { \sum_{w \in C(w_1) \cup C(w_2)} \max (P(w|w_1),P(w|w_2))}\\
\label{eq:DiceCP}
\text{\itshape Dice}^{\text{\em CP}}(w_1,w_2)& =& \frac{2 \times \sum_{w \in C(w_1) \cup C(w_2)} \min (P(w|w_1),P(w|w_2))}
                                       { \sum_{w \in C(w_1)} P(w|w_1) + \sum_{w \in C(w_2)} P(w|w_2)} \\
&=& \frac{2 \times \sum_{w \in C(w_1) \cup C(w_2)} \min (P(w|w_1),P(w|w_2))}{1 + 1} \\
\label{eq:DiceSimple}
&=& \sum_{w \in C(w_1) \cup C(w_2)} \min (P(w|w_1),P(w|w_2))
\end{eqnarray}

\noindent \aa{Observe that the special nature of the membership
function forces the Dice coefficient to equate to simplified form 
(\ref{eq:DiceSimple}) which
is also the numerator of the Jaccard coefficient. Since Dice and Jaccard
are identical in terms of ranking word pairs, use of this simplified form
is computationally optimal if one decides to use the Dice or Jaccard coefficient
with conditional probability as the strength of association.}

\aa{\inlinecite{DaganMM95} use a weighted version of the Jaccard coefficient 
on pseudo-fuzzy sets with PMI as the strength of association. They 
do not provide quantitative comparison with other distributional measures
and do not
derive their measure as shown above. Viewing co-occurrence information as pseudo-fuzzy
sets enabling the use of any of the numerous set operations to determine distributional
similarity is a novel approach. Part of our future research is to determine how well
such measures fare compared to the others.
}

\subsection{Mutual Information--Based Measures}
\inlinecite{Hindle90} was one of the first to factor the strength
of association of co-occurring words into a distributional similarity measure. The hypothesis
is that the more similar the association of co-occurring words
with the two target words, the more semantically similar they are. 
Hindle\endnote{In their respective papers, Donald Hindle and 
Dekang Lin refer to pointwise mutual information as mutual
information.} used pointwise mutual information (PMI) as the strength of association.
Consider the nouns $n_j$ and
$n_k$ that exist as objects of verb $v_i$ in different
instances within a text corpus. Hindle used formula
(\ref{eq:Hindle1}) to determine the distributional similarity of $n_j$ and $n_k$
solely from their occurrences as object of $v_i$. \aa{The minimum of
the two PMIs captures the similarity in the strength of association of $v_i$
with each of the two nouns. Note that in case of negative PMI values,
the maximum function captures the PMI which is lower in absolute value.}
\begin{equation}
\label{eq:Hindle1}
\text{\itshape Hin}_{\text{\itshape obj}} (v_i, n_j, n_k) = \left\{ \begin{array}{l} \min(I(v_i,n_j), I(v_i,n_k)),\\ \qquad \quad \text{if}\; I(v_i,n_j) > 0\; \text{and}\; I(v_i,n_k) > 0 \\
						 \mid \max(I(v_i, n_j), I(v_i, n_k))\mid,\\ \qquad \quad   \text{if}\; I(v_i, n_j) < 0\; \text{and}\; I(v_i, n_k) < 0 \\
						 0, \quad \quad \text{otherwise} \end{array} \right.
\end{equation}
\noindent $I(n, v)$ stands for the PMI (word association ratio, to be more precise)
between the words $n$ and $v$. Hindle used an analogous formula to calculate
the distributional similarity ($Hin_{subj}$) using the subject-verb relation. The overall
distributional similarity between any two nouns is calculated by the formula (\ref{eq:Hindle2}).
\begin{equation}
\label{eq:Hindle2}
\text{\itshape Hin}(n_1,n_2) = \sum_{i = 0}^{N} \left( \text{\itshape Hin}_{\text {\itshape obj}}(v_i, n_1, n_2) + \text{\itshape Hin}_{\text{\itshape subj}}(v_i, n_1, n_2) \right)
\end{equation}
\noindent The measure
gives similarity scores from 0 (maximally dissimilar) to infinity (maximally similar).
Note that in Hindle's measure, the set of co-occurring words used is restricted
to include only those words that have the same syntactic relation with both
target words (either verb-object or verb-subject). This is therefore a measure
of distributional similarity and not distributional relatedness. \bb{A form of
Hindle's measure where all co-occurring words are used, making it a measure of
distributional relatedness, is shown below:}
\begin{equation}
\label{eq:Hindle3}
\text{\itshape Hin}_{\text{\itshape{rel}}}(w_1,w_2) = \sum_{w \in C(w)} \left\{ \begin{array}{l} \min(I(w,w_1), I(w,w_2)),\\ \qquad \quad \text{if}\; I(w,w_1) > 0\; \text{and}\; I(w,w_2) > 0 \\
						 \mid \max(I(w, w_1), I(w, w_2))\mid,\\ \qquad \quad   \text{if}\; I(w, w_1) < 0\; \text{and}\; I(w, w_2) < 0 \\
						 0, \quad \quad \text{otherwise} \end{array} \right.
\end{equation}
\noindent{$C(x)$ is the set of words that co-occur with word $x$.}


\inlinecite{Lin98C} suggests a different measure derived from his
information theoretic definition of similarity \cite{Lin98B}. Further,
he uses a broad set of syntactic relations apart from subject-verb and 
verb-object relations and shows that using multiple  relations is beneficial even by
Hindle's measure. He first extracts triples of the form $(x,r,y)$ from
the partially parsed text, where the word $x$ is related to $y$ by the
syntactic relation $r$. If $I(x,r,y)$ is the information contained in
the proposition: the triple $(x,r,y)$ occurred a constant $c$ times,
then Lin defines the distributional similarity between two words, $w_1$ and $w_2$, as follows:

\begin{equation}
\label{eq:LinCorpus}
\text{\itshape Lin}(w_1,w_2) = \frac{\sum_{(r,w)\, \in\, T(w_{1})\, \cap\, T(w_{2})} \left(I(w_{1},r,w) + I(w_{2},r,w)\right)}
           {{\sum_{(r,w')\, \in\, T(w_1)} I(w_1,r,w') + \sum_{(r,w'')\, \in\, T(w_2)} I(w_2,r,w'')}}
\end{equation}

\noindent $T(x)$ is the set of all word pairs $(r,y)$ such that the pointwise mutual information
$I(x,r,y)$, is positive. Note that this is different from \inlinecite{Hindle90}
where even the cases of negative PMI were also considered. \aa{As mentioned earlier,
\inlinecite{ChurchH89} show that it is hard to accurately predict
negative word association ratios with confidence. Thus, co-occurrence pairs with negative
PMI are ignored. The measure gives similarity scores from 0 (maximally dissimilar) to 1 (maximally similar).}

Lin's measure distinguishes itself from that of Hindle in two respects.
Firstly, he normalizes the distributional similarity between two words ($w_1$ and $w_2$) determined by their
PMI with common co-occurring words by the total PMI of
$w_1$ and $w_2$ with the rest of the related words. This is a
significant improvement as now high PMI of the target words
with shared co-occurring words does not guarantee a high distributional similarity score. 
As an additional requirement, the target words must have low PMI 
with words they do not both co-occur with.
The second difference in the two formulae is that Hindle uses a minimum of
the PMI between each of the target words and the shared
co-occurring word, while Lin uses the sum. Taking
the sum has the drawback of not penalizing for a mismatch in strength
of co-occurrence, as long as $w_1$ and $w_2$ both co-occur with a word.
We suggest a new measure of distributional similarity 
(denoted by $\text{\itshape Saif}$) which counters this but keeps the
normalizing factor of Lin's measure:
\begin{equation}
\label{eq:Saif}
\text{\itshape Saif}(w_1,w_2) = \frac{2 \times \sum_{(r,w)\, \in\, T(w_{1})\, \cap\, T(w_{2})} \min(I(w_{1},r,w), I(w_{2},r,w))}
           {\sum_{(r,w') \in T(w_1)} I(w_1,r,w') + \sum_{(r,w'') \in T(w_2)} I(w_2,r,w'')}
\end{equation}
\noindent The multiplication by two is done to get scores in the range of 
0 to 1 (note that the sum in Lin's formula was replaced by a min). The 
multiplication has no effect on the relative ranking of word pairs
by their similarities. Also notice that like Hindle's measure, both Lin's
and mine are measures of distributional similarity.
\inlinecite{Hindle90} used a portion of the {\em Associated Press} news stories
(6 million words) to classify the nouns into semantically related
classes. \inlinecite{Lin98C} used his measure to generate a thesaurus
from a 64-million-word corpus of the {\em Wall Street Journal, San Jose Mercury}
and {\em AP Newswire}. He also provides a 
framework for evaluating automatically generated thesauri by comparing
them with WordNet-based and Roget-based thesauri. He shows that the thesaurus
created with his measure is closer to the WordNet and Roget-based thesauri than
that of Hindle.

\subsubsection{Mutual Information--Based Spatial and Fuzzy Metrics}

Variations of the spatial metrics (equations (\ref{eq:Cos}), (\ref{eq:L1}), and 
(\ref{eq:L2})) that use pointwise mutual information instead 
of conditional probability as the strength of association are possible.
Following are the formulae for mutual information--based spatial metrics.

\begin{eqnarray}
\label{eq:MI-based spatial metrics}
\text{\itshape Cos}^{\text {\itshape MI}}(w_1,w_2)& =& \frac{\sum_{w \in C(w_1) \cup C(w_2)} \left( I\left(w,w_1\right) \times I\left(w,w_2\right)\right)}
{\sqrt{\sum_{w \in C(w_1)} I(w,w_1)^2} \times \sqrt{\sum_{w \in C(w_2)} I(w,w_2)^2} } \\
\label{eq:L1 MI}
L_1^{\text {\itshape MI}}(w_1,w_2)& =& \sum_{w \in C(w_1) \cup C(w_2)} \mid I(w,w_1) - I(w,w_2) \mid \\
\label{eq:L2 MI}
L_2^{\text {\itshape MI}}(w_1,w_2)& =& \sqrt{\sum_{w \in C(w_1) \cup C(w_2)} (I(w,w_1) - I(w,w_2))^2}
\end{eqnarray}

\noindent Use of pointwise mutual information as the strength of association in the fuzzy
metrics (see equations (\ref{eq:DiceCP}) and (\ref{eq:JaccardCP})) discussed earlier
results in the following:

\begin{eqnarray}
\label{eq:JaccardMI}
\text{\itshape Jaccard}^{\text{\em MI}}(w_1,w_2)& =& \frac{\sum_{w \in C(w_1) \cup C(w_2)} \min (I(w,w_1),I(w,w_2))}
                                       { \sum_{w \in C(w_1) \cup C(w_2)} \max (I(w,w_1),I(w,w_2))} \\
\label{eq:DiceMI}
\text{\itshape Dice}^{\text{\em MI}}(w_1,w_2)& =& \frac{2 \times \sum_{w \in C(w_1) \cup C(w_2)} \min (I(w,w_1),I(w,w_2))}
                                       { \sum_{w \in C(w_1)} I(w,w_1) + \sum_{w \in C(w_2)} I(w,w_2)}
\end{eqnarray}

\noindent Observe that $\text{\em Saif}(w_1,w_2)$ (equation (\ref{eq:Saif})) equates to $\text{\em Dice}_{\text{\em MI}}(w_1,w_2)$ 
if the restriction to use only positive pointwise mutual information, is lifted.

\subsection{Relative Entropy--Based Measures}

\subsubsection{Kullback-Leibler divergence}

Given two probability mass functions 
$p(x)$ and $q(x)$, their {\bf relative entropy} ($D(p\Vert q)$) is:

\begin{equation}
D(p\Vert q) = \sum_{x \in X} p(x) \log \frac{p(x)}{q(x)} \hspace{1in} \text {for } q(x) \ne 0
\end{equation}

\noindent Intuitively, if $p(x)$ is the accurate probability mass function corresponding
to a random variable $X$, $D(p\Vert q)$ is the information lost on 
approximating $p(x)$ by $q(x)$. In other words, $D(p\Vert q)$ is
indicative of how different the two distributions are. Relative
entropy is also called the {\bf Kullback-Leibler divergence} or the
{\bf Kullback-Leibler distance} (denoted by $\text{\itshape KLD}$). 
 
\inlinecite{PereiraTL93} and \inlinecite{DaganPL94} point
out that words have probabilistic distributions with respect to neighboring
syntactically related words. For example, there exists a certain probabilistic
distribution ($d_1 (P(v|n_1))$, say) of a particular noun $n_1$ being the object of any verb.
This distribution can be estimated by corpus counts of parsed or chunked  text.
Let $d_2$ ($P(v|n_2)$) be the corresponding distribution for noun $n_2$. 
These distributions ($d_1$ and $d_2$) define the contexts of the two nouns ($n_1$
and $n_2$, respectively). As per the distributional hypothesis~\cite{Harris68}, 
the more these contexts are similar, the more are $n_1$ and $n_2$ semantically similar.
Thus the Kullback-Leibler distance between the two distributions
is indicative of the semantic distance between the nouns $n_1$ and $n_2$.

\begin{equation}
\begin{array}{rcll}
\text{\itshape KLD}(n_1,n_2)& =& D(d_1\Vert d_2) &  \\
                            & =& \sum_{v \in \text {Vb}} P(v|n_1) \log \frac{P(v|n_1)}{P(v|n_2)} & \text {for } P(v|n_2) \ne 0 \\
                            & =& \sum_{v \in \text {Vb'}(n_1) \cup \text {Vb'}(n_2)} P(v|n_1) \log \frac{P(v|n_1)}{P(v|n_2)} & \text {for } P(v|n_2) \ne 0 
\end{array}
\end{equation}

\noindent where $\text {\itshape Vb}$ is the set of all verbs and $\text{\itshape Vb'}(x)$
is the set of verbs that have $x$ as the object. 
The distributional similarity is determined by taking the reciprocal of the
Kullback-Leibler distance or similar suitable method.
Note that the set of co-occurring words used is restricted
to include only verbs that each have the same syntactic relation (verb-object) with both
target nouns. This is therefore a measure
of distributional similarity and not distributional relatedness.

It should be noted that the verb-object relationship is not inherent to the
measure and that one or more of any other syntactic relations may be used. 
The distributional relatedness may even be determined using all words
co-occurring with the target words. Thus a more generic expression of the
Kullback-Leibler divergence is as follows:

\begin{equation}
\begin{array}{rcll}
\label{eq:KLD}
\text{\itshape KLD}(w_1,w_2)& =& D(d_1\Vert d_2) & \\
                            & =& \sum_{w \in V} P(w|w_1) \log \frac{P(w|w_1)}{P(w|w_2)} & \text {for } P(w|w_2) \ne 0 \\
                            & =& \sum_{w \in C(w_1) \cup C(w_2)} P(w|w_1) \log \frac{P(w|w_1)}{P(w|w_2)} & \text {for } P(w|w_2) \ne 0 
\end{array}
\end{equation}

\noindent $V$ is the vocabulary (all the words found in a corpus).
$C(x)$, as mentioned earlier, is the set of words occurring (within
a certain window) with word $x$. The inverse of the distributional
distance calculated above yields the distributional relatedness of $w_1$
and $w_2$.

\pagebreak
It should be noted that the Kullback-Leibler distance is not symmetric, that
is, the distance from $w_1$ to $w_2$ is not necessarily, and even not likely,
the same as the distance from $w_2$ to $w_1$. This asymmetry is 
counter-intuitive to the general notion of semantic similarity of words, although
\inlinecite{Weeds03} has argued in favor of asymmetric measures.
Further, it is very likely that there be instances such that $P(w_1|v)$
is greater than 0 for a particular verb $v$, while due to data sparseness
or grammatical and semantic constraints,
the training data has no sentence where $v$ has the object $w_2$. This makes 
$P(w_2|v)$ equal to 0 and the ratio of the two probabilities
infinite. Kullback-Leibler divergence is not defined in such cases but approximations
may be made by considering smoothed values for the denominator.


\inlinecite{PereiraTL93} use relative entropy to create 
clusters of nouns from verb-object pairs corresponding
to a thousand most frequent nouns in the {\itshape Grolier's Encyclopedia}, June 1991
version (10 million words).
\inlinecite{DaganPL94} use Kullback-Leibler distance to estimate
the probabilities of bigrams that were not seen in a text corpus. They point
out that a significant number of possible bigrams are not seen in any
given text corpus. The probabilities of such bigrams may be determined
by taking a weighted average of the probabilities of bigrams composed
of distributionally similar words. Use of Kullback-Leibler distance as the semantic
distance metric yielded a 20\% improvement in perplexity on the {\em Wall
Street Journal} and dictation corpora provided by ARPA's HLT program~\cite{Paul91}. 

The use of distributionally similar words
to estimate unseen bigram probabilities will likely lead to erroneous results
in case of less-preferred and strongly-preferred collocations (word pairs).
\inlinecite{DianaH02} point out that even though words like {\em task} and {\em job}
are semantically very similar, the collocations they form with other words
may have varying degrees of usage. While {\em daunting task} is a 
strongly-preferred collocation, {\em daunting job} is rarely used. Thus
using the probability of one bigram to estimate that of another will
not be beneficial in such cases.

\subsubsection{$\alpha$ Skew Divergence}
{\bf $\alpha$ skew divergence} ({\em ASD}) is a slight modification of the Kullback-Leibler
divergence, that obviates the need for smoothed probabilities. It has the
following formula:
\begin{equation}
\label{eq:alpha}
\text{\itshape ASD}(w_1,w_2) = \sum_{w \in C(w_1) \cup C(w_2)} P(w|w_1) \log \frac{P(w|w_1)}{\alpha  P(w|w_2) + (1 - \alpha) P(w|w_1)}
\end{equation}
\noindent $\alpha$ is a parameter that may be varied but is usually set to $0.99$.
Note that the denominator within the logarithm is never zero with a non-zero
numerator. Also, the measure retains the asymmetric nature of the 
Kullback-Leibler divergence.

\inlinecite{Lee01} shows that $\alpha$ skew divergence performs better than 
Kullback-Leibler divergence in estimating word co-occurrence probabilities. \inlinecite{Weeds03}
achieves a correlation of \aa{$0.48$ and $0.26$ with human judgment on the Miller and Charles
word pairs using $ASD(w_1,w_2)$ and $ASD(w_2,w_1)$, respectively.}

\subsubsection{Jensen-Shannon Divergence}

A relative entropy--based measure that overcomes the drawback of asymmetry in
Kullback-Leibler divergence is the {\bf Jensen-Shannon divergence} 
a.k.a.\@ {\bf total divergence to the average} a.k.a.\@ {\bf information radius}.
It is denoted by $\text{\itshape JSD}$ and has the following formula:
\begin{eqnarray}
\label{eq:JSD}
\text{\itshape JSD}(w_1,w_2)& =& D\left(d_1 \Vert \frac{1}{2}(d_1 + d_2)\right) + D\left(d_2 \Vert \frac{1}{2}(d_1 + d_2)\right) \\ 
& =& \sum_{w \in C(w_1) \cup C(w_2)} \Bigg( P(w|w_1) \log \frac{P(w|w_1)}{\frac{1}{2}\left(P(w|w_1) + P(w|w_2)\right)} + \nonumber\\
& & \qquad \qquad P(w|w_2) \log \frac{P(w|w_2)}{\frac{1}{2}\left(P(w|w_1) + P(w|w_2)\right)} \Bigg)
\end{eqnarray}
\noindent The Jensen-Shannon divergence is the sum of Kullback-Leibler divergence between each of the individual
distributions $d_1$ and $d_2$ with the average distribution ($\frac{d_1 + d_2}{2}$).
Further, it can be shown that the Jensen-Shannon divergence avoids the problem
of zero denominator as in Kullback-Leibler divergence. The Jensen-Shannon divergence 
is therefore always well defined and, like {\bf $\alpha$ skew divergence}, obviates the need for smoothed estimates.

\aa{The Kullback-Leibler divergence, $\alpha$ Skew Divergence, and Jensen-Shannon divergence
all give distributional distance scores from 0 (maximally similar/related) to infinity (completely dissimilar/unrelated).}


\subsection{Co-occurrence Retrieval Models}

The distributional measures suggested by \inlinecite{Weeds03} are based on the notion of
substitutability. The more appropriate it is to substitute word $w_1$ 
in place of word $w_2$ in a suitable natural language task, 
the more semantically similar they are. The natural language task she focuses 
on is {\bf co-occurrence retrieval} \aa{(the retrieval of words that co-occur with a target word
from text)} and depending on the definition of
{\em appropriate} she suggests six different distributional measures 
called the {\bf co-occurrence retrieval models (CRMs)}.

Let $N_1$ be the set of co-occurrences of $w_1$ \aa{retrieved from a text corpus}
and $N_2$ that of $w_2$. In
order to determine how appropriate it is to 
substitute $w_1$ in place of $w_2$ we have to decide how important
it is to get as many co-occurrences as possible listed in $N_2$ ({\bf recall}, denoted by $R$)
and how important it is to not get co-occurrences not listed in $N_2$ ({\bf precision}, denoted by $P$).
Thus Weeds' distributional
measures have a precision component and a recall component. The final
score is a weighted sum of the precision, recall and standard $F$ measure
(see equation~(\ref{eq:CRMfinal})\endnote{$P$ is
short for $P(w_1,w_2)$, while $R$ is short for $R(w_1,w_2)$. The abbreviations
are made due to space constraints and to improve readability.}).
The weights determine the importance of precision and recall and are
determined empirically.
If precision and recall are equally important, then we get a symmetric measure
which gives the same scores to the distributional similarity of $w_1$ with $w_2$ and 
$w_2$ with $w_1$. Otherwise, we get an asymmetric measure which
assigns different similarities to the two cases. 
As substitutability is defined as a measure of distributional similarity,
metrics such as precision and recall which quantify how good the substitution is, 
are used to calculate the distributional similarity.
\begin{equation}
\label{eq:CRMfinal}
CRM(w_1,w_2) = \gamma \Biggl[ \frac{2 \times P \times R}{P + R} \Biggr]  +  (1 - \gamma) \Biggl[ \beta [ P ] + (1 - \beta) [R] \Biggr]
\end{equation}
\noindent $\gamma$ and $\beta$ are tuned parameters that lie between 0 and 1.

Weeds argues that the asymmetry in substitutability is intuitive
as in many cases
it may be okay to substitute a word, say {\em dog}, with another, say {\em animal}, but the
reverse is not likely to be acceptable as often. Since
substitutability is a measure of semantic similarity, she believes that
distributional similarity between two words should reflect this property
as well. Hence, like the Kullback-Leibler divergence, all her 
distributional similarity models are inherently asymmetric. 

A word's co-occurrence information may be specified by the set of
co-occurring words alone, or by specifying
the strength of co-occurrences, as well. This strength may be captured
by a suitable measure of word association such as
conditional probability or pointwise mutual information between the
co-occurring words and the target words. 
Also, the difference
in the strength of co-occurrence may or may not be used
to penalize the substitutability of one word for another.
\inlinecite{Weeds03} provides six distinct formulae for precision
and recall, depending on the the strength of co-occurrence
and penalty for differences in strength of association.

The precision (or recall) can be considered as the product of a 
core precision (or recall) formula (denoted by $core$) and a penalty function (denoted by $penalty$).
The CRMs that use simple counts of the common co-occurrences
in $N_1$ and $N_2$ and not the strength of associations
as core precision and recall values are called {\bf type-based
CRMs} (denoted by the superscript {\em type}). 
The CRMs that use conditional probabilities of the common co-occurrences
in $N_1$ and $N_2$ with the target words 
as core precision and recall values are called {\bf token-based
CRMs} (denoted by the superscript {\em token}). 
The CRMs that use pointwise mutual information of the common co-occurrences
in $N_1$ and $N_2$ with the target words 
as core precision and recall values are called {\bf mutual information--based
CRMs} (denoted by the superscript {\em mi}). The core precision
and recall formulae for type, token and mutual information--based
CRMs are listed below:

\begin{eqnarray}
 \text{core}_P^{type\ {}\ {}}(w_1,w_2)&     =& \frac{\mid C(w_1) \cap C(w_2) \mid}{\mid C(w_1) \mid} \\
 \text{core}_R^{type\ {}\ {}}(w_1,w_2)&        =& \frac{\mid C(w_1) \cap C(w_2) \mid}{\mid C(w_2) \mid} \\
 \text{core}_P^{token\/}(w_1,w_2)&    =& \sum_{w \in C(w_1) \cap C(w_2)} P(w|w_1) \\
 \text{core}_R^{token\/}(w_1,w_2)&       =& \sum_{w \in C(w_1) \cap C(w_2)} P(w|w_2) \\
 \text{core}_P^{mi\ {}\ {}\ {}}(w_1,w_2)&       =& \frac{\sum_{w \in C(w_1) \cap C(w_2)} I(w,w_1)}{\sum_{w \in C(w_1)} I(w,w_1)} \\
 \text{core}_R^{mi\ {}\ {}\ {}}(w_1,w_2)&          =& \frac{\sum_{w \in C(w_1) \cap C(w_2)} I(w,w_2)}{\sum_{w \in C(w_2)} I(w,w_2)} 
\end{eqnarray}

\noindent The CRMs that do not
penalize difference in strength of co-occurrence are called
{\bf additive CRMs} (denoted by the subscript {\em add}). 
The CRMs that do penalize are called {\bf difference-weighted CRMs}
(subscript {\em dw}). The penalty is 
a conditional probability--based function (\ref{eq:penalty P type}, \ref{eq:penalty R type}) for the token- and type-based
CRMs, and a mutual information--based function (\ref{eq:penalty P mi}, \ref{eq:penalty R mi}) for the mutual information--based
CRM. 
\begin{eqnarray}
\label{eq:penalty P type}
penalty_{P}^{type} = penalty_{P}^{token}& =& \frac{\min(P(w|w_1),P(w|w_2))}{P(w|w_1)} \\
\label{eq:penalty R type}
penalty_{R}^{type} = penalty_{R}^{token}& =& \frac{\min(P(w|w_1),P(w|w_2))}{P(w|w_2)} \\
\label{eq:penalty P mi}
penalty_{P}^{mi}& = &\frac{\min(I(w,w_1),I(w,w_2))}{I(w,w_1)} \\
\label{eq:penalty R mi}
penalty_{R}^{mi}& =& \frac{\min(I(w,w_1),I(w,w_2))}{I(w,w_2)} 
\end{eqnarray}
The precision and recall of additive and
difference-weighted CRMs is listed in the appendix.

\inlinecite{Weeds03} extracted verb-object pairs of 2,000 nouns from the {\em British
National Corpus (BNC)}. The verbs related to the target words
by the verb-object relation were used. Thus each of the co-occurring
verbs is related to the target nouns by the same syntactic relation
and therefore the measures capture distributional similarity,
not relatedness.
Correlation with human judgment (Miller and Charles word pairs) showed that 
difference-weighted (0.61) and additive mutual information--based measures
(0.62) performed far better than the rest of the CRMs.

\section{Discussion and Analysis of Distributional Measures}

The previous section described numerous distributional measures.
Variations of the measures are possible depending
on certain general properties of a distributional measure.
This section discusses a few of the important properties along
with an analysis of their effect in assigning semantic relatedness.

\subsection{Simple Co-occurrences vs Syntactically Related Words}

\inlinecite{Harris68}, one of the early proponents of the distributional hypothesis,
used syntactically related words to represent the context of a word.
However, the strength of association of any word appearing in the 
context of the target words may be used to determine their distributional similarity.
\inlinecite{DaganLP97}, \inlinecite{Lee99}, and \inlinecite{Weeds03}
represent the context of a noun with verbs whose object it is (single 
syntactic relation),
\inlinecite{Hindle90} represents the context of a noun with verbs with which 
it shares the verb-object or subject-verb relation, while
\inlinecite{Lin98C} uses words related to a noun by any of the many pre-decided syntactic relations
to determine distributional similarity.
\inlinecite{SchutzeP97} and \inlinecite{YoshidaYK03} use all co-occurring words in a
pre-decided window size.
Although \inlinecite{Lin98C} shows that the use of multiple syntactic
relations is more beneficial as compared to just one,
there exist no published results on whether using only syntactically
related words (as compared to all co-occurrences) improves or worsens 
the quality of semantic similarity assignment. 

Use of syntactically related words entails the
requirement of chunking or parsing the data. Once the data is
suitably parsed, the computational cost of such methods is
lower as distributional similarity is determined with much fewer words.


\subsubsection{Use of Multiple Syntactic Relations}

\inlinecite{Lin98C} used a subset of words that co-occurred with the target 
words to determine their distributional similarity. Only those co-occurrences that
are syntactically related (by any of the pre-decided list of relations) 
to the target words are chosen.
Once this restricted set of co-occurrences is determined, distributional similarity
is determined by formula~(\ref{eq:LinCorpus}) shown earlier. Observe that
the formula does not distinguish between the co-occurrences related by
different syntactic relations.
An alternative
is to calculate a distributional similarity value using each of the syntactic
relations individually and then determine the overall distributional similarity from 
these results. The overall distributional similarity may be as simple as the average similarity (see (\ref{eq:Rel_Avg}))
or the maximum (see (\ref{eq:Rel_Max})) of individual similarity results.
Distributional similarity so calculated is justified in the following 
two paragraphs, respectively.

\begin{eqnarray}
\label{eq:Rel_Avg}
\text{\itshape Sim}_{\text{\itshape Overall\_Avg}}(w_1,w_2)& =& \frac{1}{N} \big( \text{\itshape Sim}_{r1}(w_1,w_2) +
\text{\itshape Sim}_{r2}(w_1,w_2) +\nonumber\\
& & \qquad \ldots + \text{\itshape Sim}_{rN}(w_1,w_2) \big)\\
\label{eq:Rel_Max}
\text{\itshape Sim}_{\text{\itshape Overall\_Max}}(w_1,w_2)& =& \max (\text{\itshape Sim}_{r1}(w_1,w_2), 
\text{\itshape Sim}_{r2}(w_1,w_2), \nonumber\\
& & \qquad \ldots,
\text{\itshape Sim}_{rN}(w_1,w_2))
\end{eqnarray}

\noindent where $N$ is the total number of syntactic relations considered and,

\begin{equation}
\text{\itshape Sim}_{\text{\itshape ri}}(w_1,w_2) = \frac{\sum_{(ri,w)\, \in\, T(w_{1})\, \cap\, T(w_{2})} (I(w_{1},ri,w) + I(w_{2},ri,w))}
           {{\sum_{(ri,w')\, \in\, T(w_1)} I(w_1,ri,w') + \sum_{(ri,w')\, \in\, T(w_2)} I(w_2,ri,w')}}
\end{equation}

\noindent where {\em ri} is a particular syntactic relation. 

Consider the scenario where word $w'$ has a strong word association ratio
(large MI value) with $w_1$ but does not co-occur with $w_2$. The large
MI value is added to the denominator as per Lin's measure~(\ref{eq:LinCorpus}). This 
results in a low distributional similarity value. However, a number of words are considered
semantically related
even though there exist words ({\bf exclusive co-occurrences}, say) that have strong word association ratios
with one or the other target word but not both. A mark of semantically related
words is the presence of a number of 
common co-occurring words with whom they are both strongly associated. One or few
strong co-occurrences of a target word that 
do not co-occur with the other target word do not imply that the target words are
semantically unrelated.
For example, consider the rather similar pair of nouns {\em bananas} and {\em mangoes}.
The adjective {\em juicy} is likely to have a large association ratio with {\em mangoes}
but not so with {\em bananas}. 
The large MI value of {\em mangoes} and {\em juicy}
may lead to an excessively low distributional similarity value as per Lin's measure~(\ref{eq:LinCorpus}).
Averaging the different distributional similarity values (as in (\ref{eq:Rel_Avg})) calculated from individual syntactic 
relations instead of Lin's original 
method moderates the strongly negative effect of such exclusive co-occurrences
by restricting it to a particular syntactic relation (in this case, adjective-noun). 
It should be noted that the disparity in the strength of association of {\em mangoes}
and {\em juicy} versus {\em banana} and {\em juicy}, is useful in bringing
out the differences between {\em mango} and {\em banana} which may be used
to determine that {\em mango} and {\em orange} are more semantically related
than {\em mango} and {\em banana}. However, as pointed out earlier, we do not 
want a strong co-occurrence to have an adverse affect on the estimation of
distributional similarity in all other cases.

Taking the maximum of individual distributional similarity values~(\ref{eq:Rel_Max}) takes the aforementioned idea
one step ahead and is grounded in the following hypothesis: 
\begin{quote}
Different
syntactic relations are accurate predictors of the semantic similarity for different
pairs of words. 
\end{quote}
\noindent For example, fruits tend to have strong word 
associations with adjectives like {\em sweet, bitter, ripe} and {\em juicy},
and low association values with verbs that they are related to by the subject-verb
relation. For example, consider the sentences: 
\begin{center}
{\tt the ripe mango fell to the ground} \\
{\tt the ripe plum fell to the ground} 
\end{center}
\noindent The words {\em ripe} and {\em mango} are related by the adjective-noun
relation and are likely to have a large value of association. On the
other hand, {\em mango} and {\em fell} which are the subject and verb,
respectively, are likely to have a low measure of association because 
almost anything can fall.
The adjective-noun relation is thus expected to yield a higher distributional similarity value
than the subject-verb relation. Employing equation~(\ref{eq:Rel_Max})
in this case 
will mean that co-occurrences related to the target words by the adjective-noun
relation will be used to determine the distributional similarity while all other co-occurrences
will be ignored. Thus only the relation 
that has the strongest associated co-occurrences is used to determine the
distributional similarity as these co-occurrences are expected to be the best predictors
of semantic similarity.
A measure where other sets of co-occurrences,
which are weak predictors of semantic similarity, are allowed to influence the result may
cause more harm than benefit. Flipping the argument on its head,
target words predicted to be strongly distributionally similar by two
or more syntactic relations should be assigned higher distributional
similarity values than in the case of just one. Using the maximum
method will loose out on this information.

Part of our future work will be to determine if calculating individual 
similarity values from different syntactic relations and then arriving at
the final similarity is closer to human judgment or not. Also, as 
pointed out, both the average or maximum approaches have their advantages
and disadvantages. It will be interesting to determine which method
gives semantic similarity values closer to the human notion of semantic 
similarity.

\subsection{Compositionality}

The various measures of distributional similarity may be divided into two
kinds as per their composition. In certain  measures each co-occurring 
word contributes to some {\em finite calculable} distributional distance between
the target words. 
The final score of distributional distance is
the sum of these contributions. 
We will call such measures {\bf compositional measures}. 
The relative entropy--based measures,
$L_1$ norm and $L_2$ norm fall in this category. 
On the other hand,
the cosine measure along with
Hindle's and Lin's mutual information--based measures
belong to the category of what we call {\bf non-compositional} measures.
Each co-occurring word shared by both target
words contributes a score to the numerator and the denominator. 
Words that co-occur with just one of the two target words
contribute scores only to the denominator. The ratio is calculated once all
co-occurring words are considered.
Thus the distributional distance contributed
by individual co-occurrences  is not calculable and the final semantic
distance cannot be broken down into compositional distances contributed
by each of the co-occurrences. 

It must be noted that it is not clear as to which of the two kinds of measures 
(compositional or non-compositional) 
resembles human judgment more closely and how much they differ
in their ranking of word pairs. Our future work aims to determine this.

\subsubsection{Primary Compositional Measures}

The compositional measures of distributional similarity (or relatedness) capture the contribution
to distance between the target words ($w_1$ and $w_2$) due to a co-occurring word by three
primary mathematical manipulations of the co-occurrence distributions ($d_1$ and $d_2$):
the {\bf difference}, denoted by $\text{\itshape Dif}$ (as in $L_1$ norm), {\bf division}, 
denoted by $\text{\itshape Div}$ (as in the relative entropy--based 
measures) and {\bf product}, denoted by $\text{\itshape Pdt}$ (as in the conditional probability or mutual information--based cosine method). 
We will call the three types of compositional measures {\bf primary compositional
measures (PCM)}. Their form is depicted below:

\begin{eqnarray}
\label{eq:diff}
{\text{\itshape Dif}}& =& \sum_{w \in C(w_1) \cup C(w_2)} \left| P(w|w_1) - P(w|w_2) \right| \\
\label{eq:div}
{\text{\itshape Div}}& =& \sum_{w \in C(w_1) \cup C(w_2)} \left| \log \frac{P(w|w_1)}{P(w|w_2)} \right| \\
\label{eq:pdt}
{\text{\itshape Pdt}}& =& \sum_{w \in C(w_1) \cup C(w_2)} \frac{P(w|w_1) \times P(w|w_2)}{\text{\itshape Scaling Factor}} 
\end{eqnarray}

\noindent Observe that by taking absolute values in expressions (\ref{eq:diff}) and
(\ref{eq:div}), the variation in the distributions for different co-occurring words
has an additive affect and not one of cancellation. This corresponds to our
distributional hypothesis --- the more the disparity in distributions, the more is
the semantic distance between the target words. The product form (\ref{eq:pdt}) also
achieves this and is based on the theorem:
\begin{quote}
The product of any two numbers will
always be less than or equal to the square of their average. 
\end{quote}
In other words,
the more two numbers are close to each other in value, the higher is the ratio of 
their product to a suitable
scaling factor (for example, the square of their average). 
Note that the difference and division measures 
give higher values when there is large disparity between the strength 
of association of co-occurring words with the target words. They are therefore
measures of distributional distance and not distributional similarity.
The product method gives higher values when the strengths of association
are closer, and is a measure of distributional relatedness.

Although all three methods seem intuitive, each produces different distributional similarity
values and more importantly, given a set of word pairs, each is likely
to rank them differently. For example, consider the division and difference
expressions applied to word pairs ($w_1$, $w_2$) and ($w_3$, $w_4$). For simplicity, let there be
just one word $w'$ in the context of all the words. Given:

\begin{eqnarray*}
P(w'|w_1) = 0.91 \\
P(w'|w_2) = 0.80 \\ 
P(w'|w_3) = 0.60 \\ 
P(w'|w_4) = 0.50 
\end{eqnarray*}


\noindent The distributional distance between word pairs as per the difference PCM:

\begin{eqnarray*}
\text{\itshape Dif\/}(w_1, w_2)& =& | 0.91 - 0.8 | = 0.11 \\
\text{\itshape Dif\/}(w_3, w_4)& =& | 0.6 - 0.5 |  = 0.1 
\end{eqnarray*}


\noindent The distributional distance between word pairs as per the division PCM:

\begin{eqnarray*}
\text{\itshape Div}(w_1, w_2)& =& \left| \log \frac{0.91}{0.8} \right| \; = 0.056 \\
\text{\itshape Div}(w_3, w_4)& =& \left| \log \frac{0.6}{0.5} \right| \; = 0.079 
\end{eqnarray*}

\noindent Observe that for the same set of co-occurrence probabilities, the difference-based
measure ranks the ($w_3, w_4$) pair more distributionally similar (lower distributional distance), while the division-based measure
gives lower distributional similarity values for word pairs having large
co-occurrence probabilities. This behavior is not intuitive and it remains
to be seen, by experimentation, as to which of the three, difference, division
or product, yields distributional similarity measures closest to human notions
of semantic similarity.


The $L_1$ norm is a basic implementation of the difference
method. A simple product-based measure of distributional similarity is as proposed below:

\begin{equation}
\label{eq:prod}
{\text{\itshape Pdt}}^{\text{\itshape Avg}}(w_1,w_2) = \sum_{w \in C(w_1) \cup C(w_2)} \frac{P(w|w_1) \times P(w|w_2)}{(\frac{1}{2}(P(w|w_1) + P(w|w_2)))^2} 
\end{equation}

\noindent The scaling factor used is the square of the average probability.
It can be proved that if the sum of two variables is equal to a constant ($k$, say).
Their values must be equal to $k/2$ in order to get the largest product.
Now, let $k$ be equal to the sum of $P(w|w_1)/(P(w|w_1) + P(w|w_2))$ and 
$P(w|w_2)/(P(w|w_1) + P(w|w_2))$. This sum will always be equal to $1$ and hence
the product ($Z$) will be largest only when the two numbers are equal i,e, $P(w|w_1)$
is equal to $P(w|w_2)$. In other words, the farther $P(w|w_1)$ and $P(w|w_2)$
are from their average, the smaller is the product $Z$. Therefore, the measure
gives high scores for low disparity in strengths of co-occurrence and low
scores otherwise.
The incorporation of $2$ in the scaling factor results in a measure that ranges between $0$ and
$1$. 

The relative entropy--based methods use a weighted division method. 
Observe that
both Kullback-Leibler divergence  \aa{(formula repeated here for convenience --- equation (\ref{eq:KLDII}))} 
and Jensen-Shannon divergence do not take absolute
values of the division of co-occurrence probabilities. \aa{This will mean that
if $P(w|w_1) > P(w|w_2)$, 
the logarithm of their ratio will be positive
and if $P(w|w_1) < P(w|w_2)$, the logarithm will be a negative number.}
Therefore, there will be a cancellation of contributions to distributional distance by 
words that have higher co-occurrence probability with respect to $w_1$
and words that have a higher co-occurrence probability with respect to $w_2$.
\aa{Observe however that the weight $P(w|w_1)$ multiplied to the logarithm means that in general
the positive logarithm values receive higher weight than the negative ones, resulting
in a net positive score. Therefore, with no absolute value of the logarithm, as in the KLD,
the weight plays a crucial role.}
A modified Kullback-Leibler divergence ($D^{\text{\itshape Abs}}$) which incorporates 
the absolute value is suggested in equation (\ref{eq:KLDAbs}):

\begin{eqnarray}
\label{eq:KLDII}
& &{\text{\itshape KLD}}(w_1,w_2) = D(d_1\Vert d_2) = \sum_{w \in C(w_1) \cup C(w_2)} P(w|w_1) \log \frac{P(w|w_1)}{P(w|w_2)}\\
\label{eq:KLDAbs}
& &{\text{\itshape KLD}}^{\text{\itshape Abs}}(w_1,w_2) = D^{\text{\itshape Abs}}(d_1\Vert d_2) = \sum_{w \in C(w_1) \cup C(w_2)} P(w|w_1) \left| \log \frac{P(w|w_1)}{P(w|w_2)} \right|\nonumber\\
& &
\end{eqnarray}

\noindent The updated Jensen-Shannon divergence measure will remain the same as in 
equation (\ref{eq:JSD}), except that it is a manipulation of $D^{\text{\itshape Abs}}$ 
and not the original Kullback-Leibler divergence (relative entropy).

\begin{equation}
\label{eq:IRadAb}
{\text{\itshape JSD}}^{\text{\itshape Abs}}(w_1,w_2) = D^{\text{\itshape Abs}}(d_1 \Vert \frac{1}{2}(d_1 + d_2)) + D^{\text{\itshape Abs}}(d_2 \Vert \frac{1}{2}(d_1 + d_2))
\end{equation}

\noindent \aa{Note that once the absolute value of the logarithm is taken, it no longer makes
much sense to use an asymmetric weight ($P(w|w_1)$) as in the KLD or as necessary to use
a weight at all.} Equation~(\ref{eq:UnwKLD}) 
shows a simple
division-based measure. It is an
unweighted form of ${\text{\itshape KLD}}^{\text{\itshape Abs}}(w_1,w_2)$ and so we will call it 
${\text{\em KLD}}_{\text {\em Unw}}^{\text{\itshape Abs}}$.

\begin{eqnarray}
\label{eq:UnwKLD}
\text{\itshape KLD}_{\text{\itshape Unw}}^{\text{\itshape Abs}}(w_1,w_2) = \text{\itshape Div}(w_1,w_2) 
                                                   = \sum_{w \in C(w_1) \cup C(w_2)} \left| \log \frac{P(w|w_1)}{P(w|w_2)} \right| 
\end{eqnarray}

\noindent Experimental evaluation of these suggested modifications of Kullback-Leibler divergence and
informations radius is part of future work. 

\subsubsection{Weighting the PCMs}
The performance of the primary compositional measures may be improved
by adding suitable weights to the distributional distance contributed by each co-occurrence.
The idea is that some co-occurrences may be better indicators of semantic
distance than others. Usually, a formulation of the strength of association of the co-occurring
word with the target words is used as weight, the hypothesis being that
a strong co-occurrence is likely to be strong indicator of semantic distance.

Weighting the primary compositional measures results in some of the existing
measures. For example, as pointed out earlier, the Kullback-Leibler divergence
is a weighted form of the division measure (not considering the absolute
value). Here, the conditional probability of a co-occurring word with respect
to the first word ($P(w|w_1)$) is used as the weight. Since the weight is
dependent on the first word and not the other, we have asymmetry. A more
symmetric weight could be the average of the conditional probabilities
between the co-occurring word and each of the two target words. A symmetrically
weighted division PCM ${\text{\itshape Saif}}^{\text{\itshape Div}}_{\text{\itshape AvgWt}}$
is shown below:
\begin{eqnarray}
{\text{\itshape Saif}}^{\text{\itshape Div}}_{\text{\itshape AvgWt}}(w_1,w_2)& =& \sum_{w \in C(w_1) \cup C(w_2)} \frac{1}{2}\left(P(w|w_1) + P(w|w_2)\right) \left| \log \frac{P(w|w_1)}{P(w|w_2)} \right|\nonumber\\
& &
\end{eqnarray}
\noindent We can have corresponding, symmetric weighted Jensen-Shannon divergence and
$\alpha$ skew divergence. $L_2$ norm is a weighted version of the $L_1$ norm,
the weight being: $P(w|w_1) - P(w|w_2)$. 
A simple product measure with 
weights is shown below:
\begin{eqnarray}
\text{\em Pdt}_{\text{\em AvgWt}}^{\text{\em Avg}}& =& \sum_{w \in C(w_1) \cup C(w_2)} \frac{1}{2}(P(w|w_1) + P(w|w_2))
                      \frac{P(w|w_1) \times P(w|w_2)}{(\frac{1}{2}(P(w|w_1) + P(w|w_2)))^2} \nonumber\\
\label{eq:prodWtd}
 & =& \sum_{w \in C(w_1) \cup C(w_2)} \frac{P(w|w_1) \times P(w|w_2)}{\frac{1}{2}(P(w|w_1) + P(w|w_2))}
\end{eqnarray}

\aa{A better weight (which is also symmetric) may be chosen given
the following hypothesis:}
\begin{quote}
\aa{The stronger the association of a co-occurring word with a target word,
the better indicator of semantic properties of the target word it is.}
\end{quote}
\noindent \aa{The co-occurring word is likely to have different 
strengths of associations with the two target words. Taking the maximum
of the two as the weight (\inlinecite{DaganMM95}) will mean that more weight is
given to a co-occurring word if it has high strength of association with
any of the two target words. As \inlinecite{DaganMM95} point out, there is strong evidence for dissimilarity 
if the strength of association with the other target word is much lower than the maximum,
and strong evidence of similarity if the strength of association with both
target words is more or less the same. Equation~(\ref{eq:SaifWtdDiv}) is a
weighted division PCM that captures this intuition.}

\begin{eqnarray}
\label{eq:SaifWtdDiv}
& &{\text{\itshape Saif}}^{\text{\itshape Div}}_{\text{\itshape MaxWt}}(w_1,w_2)\nonumber\\
& &= \sum_{w \in C(w_1) \cup C(w_2)}
 \frac{\max \left(P(w|w_1),P(w|w_2)\right)}
               {\sum_{w' \in C(w_1) \cup C(w_2)}  \max \left(P(w'|w_1),P(w'|w_2)\right)}
	 \left| \log \frac{P(w|w_1)}{P(w|w_2)} \right|\nonumber\\
& &
\end{eqnarray}

\noindent \aa{Similarly weighted product and difference measures may
be created. Both ${\text{\itshape Saif}}^{\text{\itshape Div}}_{\text{\itshape MaxWt}}$ and
${\text{\itshape Saif}}^{\text{\itshape Div}}_{\text{\itshape AvgWt}}$
give distributional distance scores from 0 (maximally similar/related) to infinity (completely dissimilar/unrelated).}

It would be interesting to note the effect of weighting on these measures
and also to determine which weight factor is more suitable.

\subsection{Measure of Association}

As mentioned earlier, distributional measures use
the disparity in association of the target words with their co-occurring words
to determine relatedness. ~\inlinecite{Lin98C} and ~\inlinecite{Hindle90} use 
pointwise mutual information as the measure of association. The mutual information--based 
CRMs of \inlinecite{Weeds03} 
also use the same. All other measures studied in this paper
use simple conditional probability of the co-occurring words, given the
target word. It should be noted that replacing the strength of association
in a measure with another can result in a different distributional measure.
For example, the mutual information--based spatial and
fuzzy metrics discussed earlier. Lin's measure (\ref{eq:LinCorpus}) using conditional probability
(CP) is shown below:

\begin{equation}
\text{\itshape Lin}^{\text{\itshape CP}}(w_1,w_2) = \frac{\sum_{(r,w)\, \in\, T(w_1)\, \cap\, T(w_2)} (P(w|w_1) + P(w|w_2))}
           {{\sum_{(r,w')\, \in\, T(w_1)} P(w'|w_1) + \sum_{(r,w')\, \in\, T(w_2)} P(w'|w_2)}} 
\end{equation}

Of course, in case of certain measures, for example the division-based 
primary compositional measures, use of pointwise mutual information
and conditional probability is equivalent.

\begin{eqnarray}
{\text{\itshape Div}}^{{\text{\itshape MI}}}(w_1,w_2)& =& \sum_{w \in C(w_1) \cup C(w_2)} \left| \log \frac{\frac{P(w,w_1)}{P(w)P(w_1)}}{\frac{P(w,w_2)}{P(w)P(w_2)}} \right| \\
                                             & =& \sum_{w \in C(w_1) \cup C(w_2)} \left| \log \frac{P(w|w_1)}{P(w|w_2)} \right| \\
											 & =& Div(w_1,w_2)
\end{eqnarray}

\inlinecite{Weeds03} shows that her mutual information--based CRMs exhibit higher correlation
with human judgment on the Miller and Charles word pairs compared
to the ones that use conditional probability. It remains to be seen
if other measures follow the same pattern. 

\subsection{Predictors of Semantic Relatedness}
Given a pair of target words, the vocabulary may be divided into 
three sets: (1) the set of words that co-occur with both target
words (common); (2) words that co-occur with exactly one of the two target
words (exclusive); (3) words that do not co-occur with either of the two
target words. 
~\inlinecite{Hindle90} 
uses evidence only from words that co-occur with both target words
to determine the distributional similarity. All the other measures discussed in this paper
so far, use words that co-occur with just one target word, as well.

One can argue that the more there are common co-occurrences between
two words, the more they are related. For example, {\em drink} and 
{\em sip} may be considered related as they have a number of common 
co-occurrences such as {\em water, tea} and so on. Similarly,
{\em drink} and {\em chess} can be deemed unrelated as words that
co-occur with one, do not with the other. For example, {\em water}
and {\em tea} do not usually co-occur with {\em chess}, while {\em 
castle} and {\em move} are not found close to {\em drink}. Measures
that use all co-occurrences (common and exclusive) tap into
this intuitive notion.
However, certain strong exclusive co-occurrences can adversely
effect the measure. 
\aa{Consider the classic {\em strong tea}
vs {\em powerful tea} example (\inlinecite{Halliday66}).} The words {\em
strong} and {\em powerful} are semantically very related.
However, the word {\em coffee} is likely to 
co-occur with {\em strong} but not with {\em powerful}. Further,
{\em strong} and {\em coffee} can be expected to have a large
value of association as given by a suitable measure, say PMI.
This large PMI value, if used in the distributional relatedness formula, can 
greatly reduce the final value. Thus it is not clear if
the benefit of using all co-occurrences is outweighed by the
drawback pointed out.

A further advantage of using only common co-occurrences is that
the Kullback-Leibler divergence can now be used without the
need of smoothed probabilities. 

\begin{equation}
\text{\em KLD}_{\text{\em Com}}(w_1,w_2) = \sum_{w \in C(w_1) \cap C(w_2)} P(w|w_1) \log \frac{P(w|w_1)}{P(w|w_2)}
\end{equation}

\noindent Observe that we are taking the intersection of the 
set of co-occurring words instead of union as in the original
formula (\ref{eq:KLD}).

\subsection{Capitalizing on Asymmetry}
Given a hypernym-hyponym pair ({\em automobile-car}, say)
asymmetric distributional measures such as the Kullback-Leibler divergence,
$\alpha$ skew divergence and the CRMs generate different 
values as the distributional similarity of $w_1$ with $w_2$ as compared to $w_2$
with $w_1$. Usually, if $w_1$ is a more generic concept than $w_2$,
the measures find $w_1$ to be more distributionally similar to $w_2$ than
the other way round. \inlinecite{Weeds03} argues that this behavior is intuitive as
it is more often okay to substitute a generic concept in place of a specific one
than vice versa, and substitutability is a indicator of semantic similarity.
On the other hand, in most cases the notion of asymmetric semantic similarity
is counter-intuitive, and possibly detrimental. Further, in case 
two words share a hypernym-hyponym relation, they are likely to be
highly semantically similar. Thus given two words, it may make sense to always choose the 
higher of the two distributional similarity values suggested by an asymmetric measure
as the final distributional similarity between the two. This way an asymmetric measure (${\text{\em Sim}}_{\text{\em Asym}}$)
can easily be converted into a symmetric one (${\text{\em Sim}}_{\text{\em Asym}}$), 
while still capitalizing 
on the asymmetry to generate more suitable distributional similarity values for 
hypernym-hyponym word pairs. Equation~(\ref{eq:SymMax}) states the formula for the
proposed conversion. A specific implementation on the KL divergence
formula is given in equation~(\ref{eq:KLDMax})

\begin{eqnarray}
\label{eq:SymMax}
\text{\em Sim}_{\text{\em Max}}(w_1, w_2)& =& \max (Sim_{Asym}(w_1,w_2), Sim_{Asym}(w_2,w_1)) \\
\label{eq:KLDMax}
\text{\em KLD}_{\text{\em Max}}(w_1, w_2)& =& \max (\text{\em KLD}(w_1,w_2), \text{\em KLD}(w_2,w_1))
\end{eqnarray}

Another method to convert an asymmetric measure of distributional
similarity (or relatedness) into a symmetric one is by taking the average (formula~\ref{eq:SymAvg})
of the two possible similarity values. A specific implementation on the KL divergence
formula is given in equation~(\ref{eq:KLDAvg})

\begin{eqnarray}
\label{eq:SymAvg}
& &\; \; \text{\em Sim}_{\text{\em Avg}}(w_1, w_2) = \frac{1}{2} (Sim_{Asym}(w_1,w_2) + Sim_{Asym}(w_2,w_1)) \\
\label{eq:KLDAvg}
& &\text{\em KLD}_{\text{\em Avg}}(w_1, w_2) = \frac{1}{2} (\text{\em KLD}(w_1,w_2) + \text{\em KLD}(w_2,w_1)) \\
& &                                         = \frac{1}{2} \sum_{w \in C(w_1) \cup C(w_2)} \left(P(w|w_1) \log \frac{P(w|w_1)}{P(w|w_2)} + P(w|w_2) \log \frac{P(w|w_2)}{P(w|w_1)}\right) \\
& &                                         = \frac{1}{2} \sum_{w \in C(w_1) \cup C(w_2)} \left(P(w|w_1) \log \frac{P(w|w_1)}{P(w|w_2)} - P(w|w_2) \log \frac{P(w|w_1)}{P(w|w_2)} \right)\\
& &                                         = \frac{1}{2} \sum_{w \in C(w_1) \cup C(w_2)} \left(P(w|w_1) - P(w|w_2)\right) \log \frac{P(w|w_1)}{P(w|w_2)}
\end{eqnarray}

Determining the effectiveness of such conversions of existing asymmetric
measures is part of our future work.

\subsection{How CRMs Fit}

The CRMs suggested by \inlinecite{Weeds03} are the first distributional measures 
to be evaluated by comparing ranked word pairs 
with those ranked by humans (Miller and Charles word pairs).
At first glance the CRMs may look quite distinct from the rest 
of the distributional measures studied so far, owing to their
rather complex formulae and multiple optimizing parameters.
However, setting the parameters to certain standard values
equates a few of the CRMs to other measures. The difference-weighted
token-based CRM suggested by Weeds has identical values for precision and recall.
She proves that the precision (or recall) is inversely
related to the $L_1$ norm measure. This seemingly odd result of equating a distributional distance
measure with a precision (or recall) value makes sense due to the following ---
as substitutability is defined as a measure of distributional similarity,
metrics such as precision and recall which quantify how good the substitution is, 
reflect the distributional similarity and are inversely related to distributional
distance.
Thus setting $\gamma = 0$ and $\beta = 1$ or $0$,
causes the CRM to behave like the $L_1$ norm. 
Further, as 
shown below, setting $\gamma = 1$ (in other words, taking  
the $F$ measure) makes the difference-weighted
mutual information--based CRM identical to the mutual information--based
Dice coefficient (\ref{eq:DiceMI}). Following\endnotemark[\value{endnote}]
is a proof of the same.
The precision and recall of the difference-weighted MI-based CRMs are 
repeated here (equations (\ref{eq:PdwMI2}) and (\ref{eq:RdwMI2})) for convenience.

\begin{eqnarray}
\label{eq:PdwMI2}
P_{dw}^{MI}(w_1,w_2) = \frac{\sum_{w \in C(w_1) \cap C(w_2)} \min(I(w,w_1),I(w,w_2))}{\sum_{w \in C(w_1)} I(w,w_1)} \\ 
\label{eq:RdwMI2}
R_{dw}^{MI}(w_1,w_2) = \frac{\sum_{w \in C(w_1) \cap C(w_2)} \min(I(w,w_1),I(w,w_2))}{\sum_{w \in C(w_2)} I(w,w_2)} 
\end{eqnarray}

\pagebreak
\begin{theorem}
\bb{The difference-weighted mutual information--based CRM equates to the mutual information--based
Dice coefficient if its parameter $\gamma$ is set to $1$.
}
\end{theorem}

\begin{pf*}{Proof}
\begin{eqnarray*}
Sim_{dw}^{MI}(w_1,w_2)& =& \gamma \Biggl[ \frac{2 \times P \times R}{P + R} \Biggr]  +  (1 - \gamma) \Biggl[ \beta [ P ] + (1 - \beta) [R] \Biggr] \\
					  &  & \\
                      & =& 1 \Biggl[ \frac{2 \times P \times R}{P + R} \Biggr]  +  (1 - 1) \Biggl[ \beta [ P ] + (1 - \beta) [R] \Biggr] \\
					  &  & \\
					  & =& \frac{2 \times P \times R}{P + R}
\end{eqnarray*}
On substituting values for $P$ and $R$ from equations~(\ref{eq:PdwMI2}) and (\ref{eq:RdwMI2}):
\begin{eqnarray*}
& &Sim_{dw}^{MI}(w_1,w_2)\\
					  &  & \\
					  & &= \frac{2 \times \left( \frac{\sum_{w \in C(w_1) \cap C(w_2)} \min(I(w,w_1),I(w,w_2))}{\sum_{w \in C(w_1)} I(w,w_1)}\right) \times \left( \frac{\sum_{w \in C(w_1) \cap C(w_2)} \min(I(w,w_1),I(w,w_2))}{\sum_{w \in C(w_2)} I(w,w_2)}\right)}{\left(\frac{\sum_{w \in C(w_1) \cap C(w_2)} \min(I(w,w_1),I(w,w_2))}{\sum_{w \in C(w_1)} I(w,w_1)}\right) + \left( \frac{\sum_{w \in C(w_1) \cap C(w_2)} \min(I(w,w_1),I(w,w_2))}{\sum_{w \in C(w_2)} I(w,w_2)}\right)} \\
					  &  & \\
					  & &= \frac{2 \times \left( \frac{\left(\sum_{w \in C(w_1) \cap C(w_2)} \min(I(w,w_1),I(w,w_2)) \right)^2}   {\left(\sum_{w \in C(w_1)} I(w,w_1)\right)\left(\sum_{w \in C(w_2)} I(w,w_2)\right)}  \right)}       {\frac{\left( \sum_{w \in C(w_1) \cap C(w_2)} \min(I(w,w_1),I(w,w_2)) \right) \left( \sum_{w \in C(w_1)} I(w,w_1) + \sum_{w \in C(w_2)} I(w,w_2)     \right)}  {\left(\sum_{w \in C(w_1)} I(w,w_1)\right)\left(\sum_{w \in C(w_2)} I(w,w_2)\right)}} \\
					  &  & \\
                      & &= \frac{2 \times \sum_{w \in C(w_1) \cap C(w_2)} \min(I(w,w_1),I(w,w_2)}   {\sum_{w \in C(w_1)} I(w,w_1) + \sum_{w \in C(w_2)} I(w,w_2)} \\
					  &  & \\
					  & &= \text{\itshape Dice}^{\text{\em MI}}(w_1,w_2) 
\end{eqnarray*}
\qed
\end{pf*}

\subsection{Hit and Miss Co-occurrences}
\aa{Lastly, we examine two kinds of co-occurrences that pose a challenge 
to existing distributional measures: (1) Word pairs that occur together
less number of times than what would be expected by chance.
Measures like PMI cannot predict their association values with confidence
and as pointed out earlier this is countered by ignoring them completely.
This means that the system misses out on evidence from 
this set of co-occurrence pairs. (2) Co-occurrence pairs formed by a
word with target words that are near synonyms.
\inlinecite{DianaH02}
point out that 
near synonyms (for example, {\itshape hidden} and {\itshape concealed})
may form strong and anti-collocations, respectively, with the same co-occurring word
(for example,
{\itshape agenda}). All distributional measures that use strength of 
association to determine semantic relatedness will consider the large
discrepancy in strength of association as evidence of unrelatedness.
Therefore, these co-occurrence
pairs, which are not ignored (unlike the previous ones),
will negatively impact the ability of distributional measures to predict semantic
relatedness of near synonyms.
It should be noted that we cannot eliminate such co-occurrences in a 
straightforward manner simply because we are not aware apriori if  the
target words are near synonyms.
It would be interesting to determine 
the precise quantitative effect of such co-occurrences on the performance
of distributional measures.
}

\subsection{Summarizing the Distributional Measures}
In the last two sections we have seen numerous distributional measures. 
Tables~\ref{tab:distributional measures of distance - I}, \ref{tab:distributional measures of distance - II},
\ref{tab:distributional measures of similarity - I}, and \ref{tab:distributional measures of similarity - II} 
listed in the appendix
summarize their properties. 

\section{Semantic Network and Ontology-Based Measures}

Creation of electronically available ontologies and semantic
networks like WordNet has allowed their use to help solve numerous 
natural language problems including the measurement of semantic
distance between two words. 
\inlinecite{Budanitsky99},
\inlinecite{BudanitskyH00} and \inlinecite{PatwardhanBT03}
have done an extensive survey of the various WordNet-based measures,
their
comparisons with human judgment on selected word pairs, and
their efficacy in applications such as spelling correction
and word sense disambiguation. Hence, this paper provides
just a brief summary of the major WordNet-based measures
of similarity and focuses on their comparison with 
distributional ones. 

One of the earliest and simplest measures is the \inlinecite{RadaMBB89}
{\bf edge counting} method. The shortest path in the network 
between the two target words ({\bf target path}) is determined. 
The more edges there are
between two words, the more distant they are. Elegant as it 
may be, the measure 
relies on the unlikely assumption that all the network edges
correspond to identical semantic distance between the nodes
they connect. Nodes in a network may be connected by numerous
relations such as hyponymy, meronymy and so on. 
Edge counts apart, \inlinecite{HirstS98} take into account
the fact that if the target path consists of edges that belong
to a number of such relations, the target words are likely
more distant. The idea is that if we start from a particular node
({\bf base word})
and take a path via a particular relation (say, hyponymy),
to a certain extent the words reached will be quite related
to the base word. However, if during the way we take edges
belonging to different relations (other than hyponymy), very 
soon we may reach words that are unrelated. Hirst and St-Onge's
measure of semantic relatedness is listed below:

\begin{equation}
\text{\itshape HS}(c_1,c_2) = C - \text{\itshape path length} - k \times d
\end{equation}

\noindent where $c_1$ and $c_2$ are the target concepts/words.
And, $d$ is the number of times an edge corresponding to a different 
relation than that of the preceding edge is taken. 
$C$ and $k$ are empirically determined constants. 

\inlinecite{LeacockC98}
used just one relation (hyponymy) and modified the path length
formula to reflect the fact that edges lower down in the {\em is-a}
hierarchy correspond to smaller semantic distance than the ones
higher up. For example, {\em sports car} and {\em car} (low in the hierarchy)
are much
more similar than {\em transport} and {\em instrumentation} (higher
up in the hierarchy) even
though both pairs of words are separated by exactly one edge in the
{\em is-a} hierarchy of WordNet. 

\begin{equation}
{\text{\itshape LC}}(c_1,c_2) =  -\log \frac{{\text{\itshape len}}(c_1,c_2)}{2D}
\end{equation}

\noindent where $D$ is the depth in the taxonomy.

\inlinecite{Resnik95} suggested a measure that used corpus statistics
along with the knowledge obtained from a semantic network.
The measure is based on the notion that the semantic similarity of two 
words may be determined from the word that represents their
similarity (the {\bf lowest common subsumer} or {\bf lowest super-ordinate (lso)}). 
The more the information
contained in this node, the more similar the two words are.
Observe that using information content (IC) has the effect of inherently 
scaling the semantic similarity measure by depth of the taxonomy. Usually, the lower
the lowest super-ordinate, the lower is the probability of occurrence of
the lso and the concepts subsumed by it, and hence, the higher is its 
information content. 

\begin{equation}
{\text{\itshape Res}}(c_1,c_2) =  -\log {\text{\itshape p}}(lso(c_1,c_2))
\end{equation}

\noindent As per the formula, given a particular lowest super-ordinate, the exact positions
of the target words below it in the hierarchy do not have any effect
on the semantic similarity. Intuitively, we would expect that word pairs closer
to the lso are more similar than those that are distant. \inlinecite{JiangC97} and \inlinecite{Lin97}
incorporate this notion into their measures which 
are arithmetic variations of the same terms.
\aa{The \inlinecite{JiangC97} measure (denoted by {\itshape JC}\/) determines how dissimilar each target concept
is from the lso ($IC(c_1) - IC(lso)$ and $IC(c_2) - IC(lso)$). The final semantic 
distance between the two concepts is
then taken to be the sum of these differences (see~\inlinecite{Budanitsky99} for more
details). 
\inlinecite{Lin97} points out that the lso is what is common between the two 
target concepts and that its information content is the 
common information between the two concepts. Lin's formula (denoted by {\itshape Lin}\/) can thus be thought of
as taking the Dice coefficient of the information in the two concepts.
}

\begin{eqnarray}
{\text{\itshape JC}}(c_1,c_2)& =&  2\log p({\text{\itshape lso}}(c_1,c_2))
- (\log(p(c_1)) + (\log(p(c_2))) \\
{\text{\itshape Lin}}(c_1,c_2)& =&  \frac{2 \times \log p({\text{\itshape lso}}(c_1,c_2))}
{\log(p(c_1)) + (\log(p(c_2))}
\end{eqnarray}

\noindent \inlinecite{BudanitskyH00} show that the Jiang-Conrath measure
has the highest correlation (0.850) with the Miller and Charles word pairs
and performs better than all other measures considered in a spelling correction task. \inlinecite{PatwardhanBT03}
get similar results using the measure for word sense disambiguation (especially of nouns).


\section{Comparison of Distributional and Ontology-Based Measures}

Distributional and ontology-based measures use distinct sources
of knowledge to achieve the same goal ---  the ability to mimic
human judgment of semantic relatedness. Owing to the difference in 
methodology, many interesting comparisons may be made. The 
next few subsections aim at bringing them to light.

\subsection{Knowledge Source versus Similarity Measure}

Ontologies are much more expensive resources than raw data, which
is freely available. Creating an ontology requires human experts,
is time intensive and rather brittle to changes in language. 
Once created, updating an ontology is again expensive and there
is usually a lag between the current state of language usage/comprehension
and the semantic network representing it. Further, the complexity
of human languages makes creation of even a near perfect semantic 
network of its concepts impossible. Thus in many ways the 
ontology-based measures are as good as the networks on which
they are based. On the other hand, large corpora, trillions of
words in size, may now be collected by a simple web crawler.
Large corpora of more formal writing are also available (for
example, the {\em Wall Street Journal} or the {\em American Printing
House for the Blind (APHB)} corpus). Therefore,
using an appropriate distributional measure that best captures
the semantic similarity--predicting information, plays a much more vital role
in case of distributional measures. 

As ontologies are a rich source of information where the various
concepts are linked together by powerful relations such as
hyponymy and meronymy, the ontological measures likely
correctly identify target words related by edges
that belong to just one relation as very similar. However,
data sparseness may force distributional measures to assign
low similarity values to clearly related word pairs. Assigning
appropriate semantic similarity values when target words are connected by
different relational edges poses a major challenge to 
ontological measures. 

\subsection{Domain-Specific Semantic Similarity}

So far, this paper has talked about {\bf universal similarity measures}.
Given a word pair, the measures each give just one similarity value.
However, two words may be very semantically similar in a certain domain but
not so much in another. For example, the word pair {\em space} and {\em time}
are closely related in the domain of quantum mechanics but not so much
in most others. Ontologies have been made for specific domains, which
may be used to determine semantic similarity specific to these domains. However,
the number of such ontologies is very limited. On the other hand, large amounts of
corpora specific to particular domains are much easier to collect, 
allowing a widespread use of distributional {\em domain-specific} similarity.

\subsection{Associated Words}

Certain word pairs have a special relation with each other. For example, {\em
strawberry} and {\em cream}, {\em doctor} and {\em scalpel}, and so on.
These words are not similar physically or in 
properties, but {\em strawberries} are usually eaten with {\em cream} and
a {\em doctor} uses a {\em scalpel} to make an incision. An ontology-based
measure will correctly identify the amount of
semantic relatedness only if such relations are inherent in the ontology.
For example, if the agent-instrument relation does not link
concepts in a semantic network (as in WordNet), the ontology-based 
measures will not identify {\em doctor} and
{\em scalpel} as related. 

Of the various distributional measures discussed, the ones
that use simple co-occurrences capture such 
semantic relatedness, as words that tend to occur together are likely 
to have large set of common co-occurring words. Measures (e.g., \inlinecite{Lin98C}, \inlinecite{Hindle90})
that consider a word $w$ to be a shared co-occurrence only if $w$
is related to both target words by the same syntactic relation,
will not find such words related, simply because such words
that tend to occur in the same sentence are likely to have
different thematic roles and thus different syntactic relations with 
common co-occurring words. 


\subsection{Multi-faceted Concepts}

The various senses of a word represent distinct concepts. Each of 
these concepts can usually be described by a number of attributes
or features. These attributes may be physical descriptions like
color, shape and composition or function, purpose and role.
Two words are adjudged similar if they share a number of such
attributes and if the strength of the shared attributes is high. 
By strength we mean
how strongly an attribute helps define the words. The more
prominent a shared feature, the more similar the two words are.
Further, it is possible that words $w_1$ and $w_2$ are related
as they share a certain set of attributes, while $w_2$ and $w_3$
are related because they share a different set of attributes.
Thus $w_1$ and $w_3$ are likely not as related as $w_1$ and $w_2$,
or $w_2$ and $w_3$. For example, the physical {\em key} is closely related
to the abstract {\em password}, as they are both {\em means of getting access}.
{\em Password} is closely related to {\em encryption} as they 
both pertain to {\em data security}.
However, the physical {\em key} has little to do with {\em encryption}
and the two are not so much related. Thus semantic relatedness is not 
necessarily transitive and may be a function of a subset of relevant
attributes, not necessarily all. 

\begin{figure}
\centerline{\includegraphics[height=4.5cm, width=12cm]{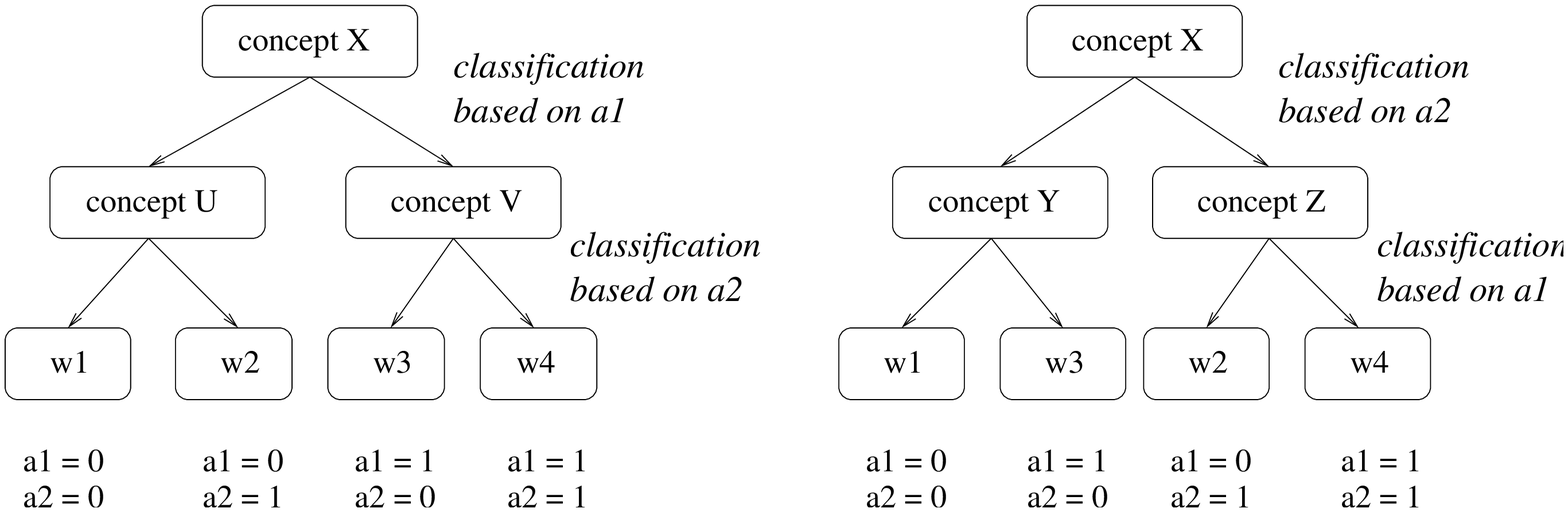}}
\caption{Hierarchy Variations.}
\label{fig:hierarchy}
\end{figure}

Hierarchies in an ontology are built by repetitive division of
concepts as per their attributes. The order in which these
attributes are used to create the tree structure can result
in dramatically different hierarchies. For example, consider
a scenario depicted in figure \ref{fig:hierarchy}, where the attributes 
$a_1$ and $a_2$ are used in 
different orders to create different hierarchies of the words 
$w_1, w_2, w_3$ and $w_4$. Notice that while $w_1$ and
$w_2$ are closer to each other than $w_1$ and $w_3$ in
hierarchy-1, it is the other way round in hierarchy-2.
Thus variations in the order of use of attributes for
creating the hierarchy can result in different sets of 
words being close to each other. 

It should be noted that
real-world semantic hierarchies are created by well formed
methodologies and hence the order of attributes used
to create the hierarchy is not arbitrary. That said, there
is room for variation and further, once a particular 
hierarchy is chosen, it captures certain semantic relations 
in its structure, while others are lost.


\begin{figure}[htbp]
\begin{center}
\includegraphics[height=5cm, width=10cm]{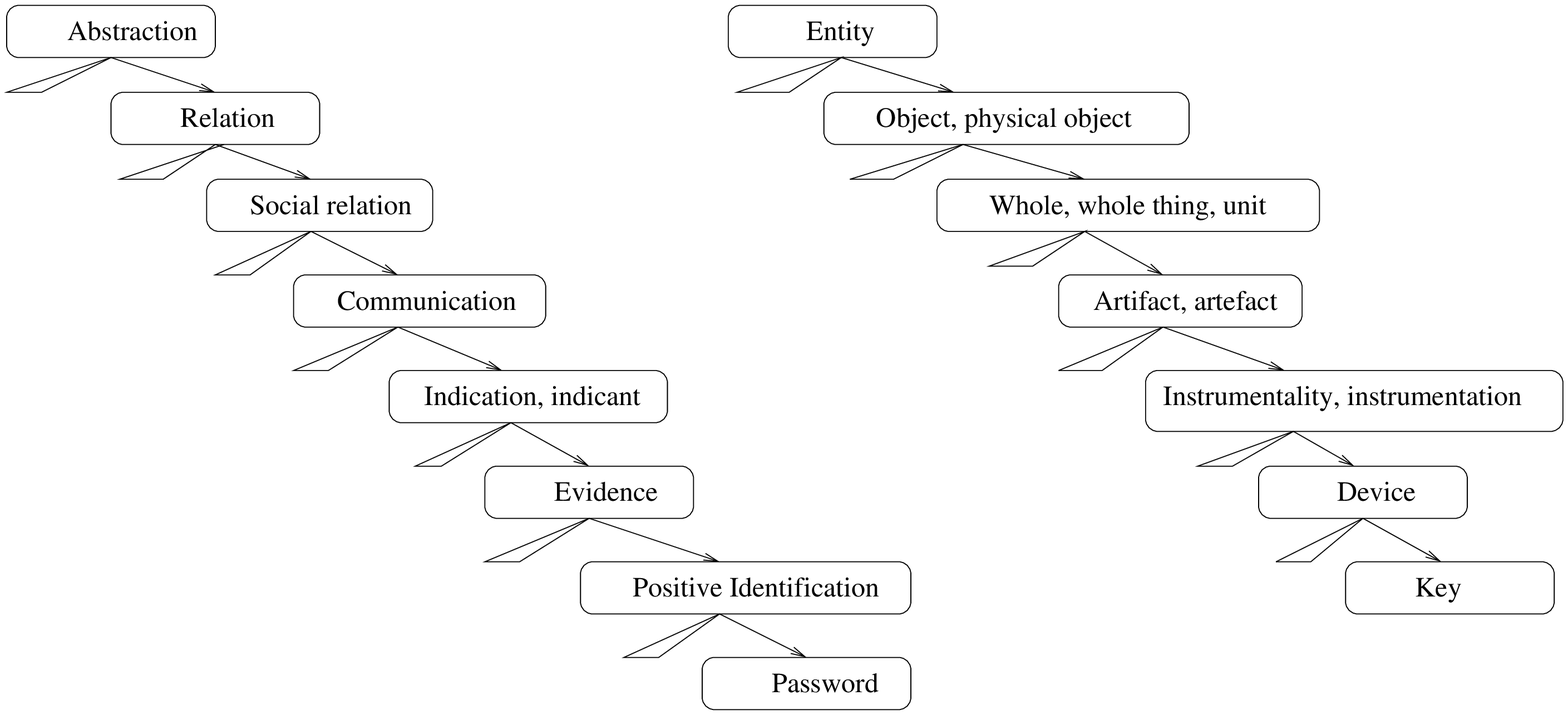}
\end{center}
\caption{{\em key} and {\em password} in the `is-a' hierarchy of WordNet}
\label{fig:WNet}
\end{figure}

In general, ontology-based measures of similarity
capitalize on the property that words that occur close to 
each other in a hierarchy share a lot of attributes and are 
therefore similar. 
However, they usually rely on a fixed hierarchy.
Word pairs that would be closer in variations of the
hierarchy are not considered. Thus ontology-based measures
are likely to wrongly assign a low semantic similarity value to
such word pairs. For example, consider {\em key} and
{\em password}. They are both {\em means to gain access}
but the is-a hierarchy in WordNet lists them in completely
different branches of the network (figure~\ref{fig:WNet}). 
The attribute determining whether the word refers to a
physical entity or an abstraction is used first to classify
the words and hence {\em key} and {\em password} fall into
different branches at the top of the hierarchy itself.
Thus an ontology-based measure is likely to find them unrelated.
Distributional measures are not bound by a fixed hierarchy
and have a better chance at appropriately identifying the
semantic similarity of such word pairs. It would be worth determining
the extent to which this is true. 

\subsection{Evaluation and Complementarity}

Ontology-based and distributional measures of similarity
have each been individually shown to be reasonable quantifiers of 
semantic similarity.  WordNet-based
measures have been used for applications such as spelling correction
and word sense disambiguation, while distributional measures have
primarily been used for estimating probabilities of unseen bigrams.
Exhaustive comparisons of 
WordNet-based measures with each other (e.g., \inlinecite{Budanitsky99},
\inlinecite{BudanitskyH00} and \inlinecite{PatwardhanBT03}) have found that
the Jiang-Conrath measure performs better than the rest. 

\inlinecite{DaganPL94} perform experiments with a few relative entropy--based 
measures and find that Jensen-Shannon divergence 
is slightly better than Kullback-Leibler divergence and $L_1$ norm in
estimating bigram probabilities of unseen words and in a pseudo--word 
sense disambiguation experiment. 
However, the various distributional measures have not been used
to rank the Miller-Charles or Rubenstein-Goodenough word pairs,
for which estimates of human judgment of semantic relatedness are available.
Experiments to this end will also enable a comparison of the 
distributional measures with the ontology-based measures
for which this data is already available.
Similar to the case of various ontological
measures, it is worth determining which distributional measure 
is closest to human notion of semantic similarity
and how well the distributional measures, which rely just
on raw data, fare against, the more expensive and knowledge rich, ontology-based
measures. 

Since the two kinds of measures rely on different knowledge
sources, there is a likelihood that distributional measures
more accurately identify the semantic similarity of a certain subset of word pairs,
while the ontological methods do so for a different subset. One of
the more important problems in the field is to quantify this
complementarity. It should be noted that since a similarity measure is 
evaluated by comparison of ranked word 
pairs and not by the similarity values alone, capitalizing on the
complementarity by creating a combined semantic similarity predictor is a 
much harder problem. 



\section{Conclusions}

The paper has provided a detailed analysis of various corpus-based
distributional measures and compared them with
measures based on ontologies and semantic networks. Merits
and limitations of the various measures were listed. New measures
that are likely to overcome the drawbacks of present distributional
measures were suggested. Specifically, a distributional measure that
keeps the best of \inlinecite{Hindle90} and \inlinecite{Lin98C}, overcoming
their respective drawbacks, was proposed. Variations of 
Kullback-Leibler divergence and Jensen-Shannon divergence that better
capture the disparity in co-occurrence probabilities were
suggested. A simple technique to convert asymmetric measures into
symmetric ones was suggested. Novel approaches are described to determine 
distributional similarity by better utilization of co-occurring words related by different 
syntactic relations. 

The paper identified significant research problems that need to
be answered through experimentation. This will help better understand how statistics
from raw data may be manipulated to determine appropriate
similarity values between two words. For example, whether the use of syntactically
related, rather than plain co-occurrences, significantly improves the 
measure? Or, are simple co-occurrences just as useful? 
What kinds of co-occurrences (common, or exclusive, as well) 
should be used to determine distributional relatedness?
Is pointwise mutual information or conditional probability
a more suitable measure of association to be used in the various distributional 
measures?
Do compositional or non-compositional distributional measures
produce more intuitive semantic similarity
values?
Which mathematical operation (difference, division, or
product) of the co-occurrence distributions yields 
values that are closest to human judgment, in case of compositional
measures? A direct evaluation of the distributional measures (other than
$L_1$ norm, $\alpha$ skew divergence and the CRMs for
which these results exist) by their
correlation with the Miller-Charles and Rubenstein-Goodenough word
pairs will provide better insight into their relative abilities
and will enable a comparison with WordNet-based measures for which the
correlation coefficients are already available. 

Lastly, the paper pointed out that even though ontological measures
are likely to perform better as they rely on a much richer knowledge
source, distributional measures have certain distinct advantages.
For example, they can easily provide domain-specific similarity values
for a large number of domains, their ability to determine similarity
of contextually associated word pairs more appropriately,
and the flexibility to identify multi-faceted concepts as
related from appropriate commonalities that may not be
explicitly encoded in a semantic network. Thus it is very 
likely that ontological measures are better at predicting 
semantic similarity for certain word pairs, while the distributional 
measures do so for others. To identify the extent of this
complementarity and a suitable combined
methodology to assign semantic similarity, remain significant problems
in the field. 
A significant challenge in achieving this is how to reconcile
the nature of the two kinds of measures --- while ontological
measures predict the semantic similarity of two concepts (or word senses),
distributional measures do so for two words. One of the problems
we intend to pursue is the development of a methodology that enables
the use of distributional measures to predict semantic similarity
of concepts, with no or little sense-tagged data. 

\appendix
\subsection{Co-occurrence Retrieval Models}
The precision and recall of additive and
difference-weighted CRMs~\cite{Weeds03}.

\begin{eqnarray}
& P_{add}^{type}(w_1,w_2) =& \frac{\mid C(w_1) \cap C(w_2) \mid}{\mid C(w_1) \mid} \\
& R_{add}^{type}(w_1,w_2) =& \frac{\mid C(w_1) \cap C(w_2) \mid}{\mid C(w_2) \mid} \\
& P_{dw}^{type}(w_1,w_2) =& \frac{\sum_{\mid C(w_1) \cap C(w_2) \mid} \frac{\min(P(w|w_1),P(w|w_2))}{P(w|w_1)}}{\mid C(w_1) \mid} \\
& R_{dw}^{type}(w_1,w_2) =& \frac{\sum_{\mid C(w_1) \cap C(w_2) \mid} \frac{\min(P(w|w_1),P(w|w_2))}{P(w|w_2)}}{\mid C(w_2) \mid}
\end{eqnarray}

\begin{eqnarray}
P_{add}^{token}(w_1,w_2)& =& \sum_{w \in C(w_1) \cap C(w_2)} P(w|w_1) \\
R_{add}^{token}(w_1,w_2)& =& \sum_{w \in C(w_1) \cap C(w_2)} P(w|w_2) \\
P_{dw}^{token}(w_1,w_2)& =& \sum_{w \in C(w_1) \cap C(w_2)} \min(P(w|w_1),P(w|w_2)) \\
R_{dw}^{token}(w_1,w_2)& =& \sum_{w \in C(w_1) \cap C(w_2)} \min(P(w|w_2),P(w|w_1))
\end{eqnarray}

\begin{eqnarray}
P_{add}^{mi}(w_1,w_2)& =& \frac{\sum_{w \in C(w_1) \cap C(w_2)} I(w,w_1)}{\sum_{w \in C(w_1)} I(w,w_1)} \\
R_{add}^{mi}(w_1,w_2)& =& \frac{\sum_{w \in C(w_1) \cap C(w_2)} I(w,w_2)}{\sum_{w \in C(w_2)} I(w,w_2)} \\
P_{dw}^{mi}(w_1,w_2)& =& \frac{\sum_{w \in C(w_1) \cap C(w_2)} \min(I(w,w_1),I(w,w_2))}{\sum_{w \in C(w_1)} I(w,w_1)} \\
R_{dw}^{mi}(w_1,w_2)& =& \frac{\sum_{w \in C(w_1) \cap C(w_2)} \min(I(w,w_1),I(w,w_2))}{\sum_{w \in C(w_2)} I(w,w_2)}
\end{eqnarray}

\noindent where $C(x)$ is the set of all co-occurrences of word $x$.
Note that in case of the difference-weighted token and mutual information--based precision
and recall formulae, there is a cancellation of a pair of terms obtained
from the core formulae and the penalty.

\subsection{Summarizing Tables}

Tables~\ref{tab:distributional measures of distance - I} and \ref{tab:distributional measures of distance - II}
list the measures of
distributional distance while 
tables~\ref{tab:distributional measures of similarity - I} and \ref{tab:distributional measures of similarity - II} 
list the
measures of distributional relatedness/similarity. \aa{If a measure is placed in a distributional distance table, it means that the 
intuition behind the measure lead to its original conception as a distance measure
and similarly for a relatedness measure.
It should be noted however that a distance measure may be converted
into a relatedness measure by taking the inverse or other such mathematical manipulation,
and vice versa.}
Apart from the formula, the tables show whether the measure is
compositional (Comp.) or not, and if so then the kind of primary compositional
measure (PCM) it is derived from.
The last column (Str.) indicates the particular strength of association
used (most commonly) in the measure --- conditional probability (CP) or pointwise mutual information (PMI).

\begin{landscape}
\begin{center}
\begin{table*}[htbp]
        \caption[Measures of distributional distance]{Measures of distributional distance.} 
        \label{tab:distributional measures of distance - I}
        \hspace{-0.05in}
		\begin{tabular}{|l c c l c c|} \hline
           {\bf Measure}    &{\bf Comp.}    &{\bf PCM}    &{\bf Formula}    &{\bf Sym.}    &{\bf Str.}\\ \hline \hline 

		                    &               &       &                                                                                                                                    &    &    \\
           ${\text{\itshape Dif}}\;\; \text{or}\;\; L_1 $                                    &$\checked$     &diff.                                              &$ \sum_{w \in C(w_1) \cup C(w_2)} \mid P(w|w_1) - P(w|w_2) \mid $                       &$\checked$         &CP\\
		                    &               &       &                                                                                                                                    &    &    \\
           $L_2$                                    &$\checked$     &diff.                          &$\sqrt{\sum_{w \in C(w_1) \cup C(w_2)} \left(P\left(w|w_1\right) - P\left(w|w_2\right)\right)^2 }$      &$\checked$         &CP\\ 
		                    &                &       &                                                                                                                                    &    &    \\
           $\text{\em KLD}$                          &$\checked$     &div.                           &$\sum_{w \in C(w_1) \cup C(w_2)} P(w|w_1) \log \frac{P(w|w_1)}{P(w|w_2)}$                                            &X                  &CP\\
		                    &                &       &                                                                                                                                    &    &    \\
           $\text{\em KLD}_{\text{\em Com}}$                          &$\checked$     &div.          &$\sum_{w \in C(w_1) \cap C(w_2)} P(w|w_1) \log \frac{P(w|w_1)}{P(w|w_2)}$                                            &X                  &CP\\
		                    &                &       &                                                                                                                                    &    &    \\
           $\text{\em KLD}^{\text{\em Abs}}$         &$\checked$     &div.                            &$\sum_{w \in C(w_1) \cup C(w_2)} P(w|w_1) \left| \log \frac{P(w|w_1)}{P(w|w_2)} \right|$                 &X                  &CP\\
		                    &                &       &                                                                                                                                    &    &    \\
           ${\text{\itshape Div}}\;\; \text{or}\;\; \text{\em KLD}_{\text{\em Unw}}^{\text{\em Abs}}$         &$\checked$     &div.                                     &$\sum_{w \in C(w_1) \cup C(w_2)} \left| \log \frac{P(w|w_1)}{P(w|w_2)}\right|$                                                     &$\checked$         &CP\\
		                    &                &       &                                                                                                                                    &    &    \\ \hline
        \end{tabular}
        \hspace{-0.20in}
        \end{table*}
\end{center}
\end{landscape}
\begin{landscape}
\begin{center}
\begin{table*}[htbp]
        \caption[Measures of distributional distance]{Measures of distributional distance (\itshape{continued}\/).} 
        \label{tab:distributional measures of distance - II}
        \hspace{-0.05in}
		\begin{tabular}{|l c c l c c|} \hline
           {\bf Measure}    &{\bf Comp.}    &{\bf PCM}    &{\bf Formula}    &{\bf Sym.}    &{\bf Str.}\\ \hline \hline 

		                    &               &       &                                                                                                                                    &    &    \\
           $\text{\em Saif}^{\text{\em Div}}_{\text{\em AvgWt}}$         &$\checked$   &div.      &$\sum_{w \in C(w_1) \cup C(w_2)} \frac{1}{2}\left(P(w|w_1) + P(w|w_2)\right) \left| \log \frac{P(w|w_1)}{P(w|w_2)} \right|$           &$\checked$         &CP\\
		                    &                &       &                                                                                                                                    &    &    \\
           $\text{\em Saif}^{\text{\em Div}}_{\text{\em MaxWt}}$         &$\checked$   &div.    
               &$\sum_{w \in C(w_1) \cup C(w_2)} \frac{\max \left(P(w|w_1),P(w|w_2)\right)}{\sum_{w' \in C(w_1) \cup C(w_2)}  \max \left(P(w'|w_1),P(w'|w_2)\right)}\times$
			   &$\checked$         &CP\\
		                    &                &                                                   &$\qquad \qquad \left| \log \frac{P(w|w_1)}{P(w|w_2)} \right|$                                                                                        &    &    \\
		                    &                &                                                   &                                                                                        &    &    \\
           $\text{\em KLD}_{\text{\em Avg}}$         &$\checked$     &div.                             &$\frac{1}{2} \sum_{w \in C(w_1) \cup C(w_2)} \left(P(w|w_1) - P(w|w_2)\right) \log \frac{P(w|w_1)}{P(w|w_2)} $   &$\checked$         &CP\\
		                    &                &                                                   &                                                                                        &    &    \\
           $\text{\em KLD}_{\text{\em Max}}$         &$\checked$     &div.                           &$\max (\text{\em KLD}(w_1,w_2), \text{\em KLD}(w_2,w_1))$                                                             &$\checked$         &CP\\
		                    &                &                                                   &                                                                                        &    &    \\

           {\em ASD}                                 &$\checked$     &div.                                        &$\sum_{w \in C(w_1) \cup C(w_2)} P(w|w_1) \log \frac{P(w|w_1)}{\alpha  P(w|w_2) + (1 - \alpha) P(w|w_1)}$            &X      &CP\\
		                    &                &                                                   &                                                                                        &    &    \\

           {\em JSD}                                 &$\checked$     &div.                                         &$\sum_{w \in C(w_1) \cup C(w_2)} \Bigl( P(w|w_1) \log \frac{P(w|w_1)}{\frac{1}{2}\left(P(w|w_1) + P(w|w_2)\right)} +$
                    &$\checked$         &CP\\
                            &                &                                                   &$\qquad \qquad P(w|w_2) \log \frac{P(w|w_2)}{\frac{1}{2}\left(P(w|w_1) + P(w|w_2)\right)} \Bigr)$                    &                   & \\ 
		                    &                &                                                   &                                                                                        &    &    \\ \hline
        \end{tabular}
        \hspace{-0.20in}
        \end{table*}
\end{center}
\end{landscape}
\begin{landscape}
\begin{center}
\begin{table*}[htbp]
        \caption[Measures of distributional relatedness/similarity - I]{Measures of distributional relatedness/similarity.}
        \label{tab:distributional measures of similarity - I}
        \hspace{-0.05in}
        \begin{minipage}{\textwidth}
        \begin{tabular}{|l c c l c c|} \hline
           {\bf Measure}    &{\bf Comp.}    &{\bf PCM}        &{\bf Formula}    &{\bf Sym.}    &{\bf Str.}\\ \hline \hline 

		                    &               &                                                   &                                                                                        &    &    \\
           $\text{\itshape{Pdt}}^{\text{\itshape{Avg}}}$        &$\checked$     &pdt.                                               &$\sum_{w \in C(w_1) \cup C(w_2)} \frac{P(w|w_1) \times P(w|w_2)}{(\frac{1}{2}(P(w|w_1) + P(w|w_2)))^2}$ &$\checked$         &CP\\
		                    &               &                                                   &                                                                                        &    &    \\
           ${\text{\itshape Pdt}}_{\text{\itshape{AvgWt}}}^{\text{\itshape Avg}}$     &$\checked$     &pdt.        &$\sum_{w \in C(w_1) \cup C(w_2)} \frac{P(w|w_1) \times P(w|w_2)}{\frac{1}{2}(P(w|w_1) + P(w|w_2))}$     &$\checked$         &CP\\ 
                            &               &                                                    &                                                &                   & \\ 
           $\cos$           &X              &n.a.\footnote[0]{{\em Note}: `n.a.' stands for `not applicable'. For example, the cos measure is not a compositional measure and therefore the type of PCM is not applicable.}
                                           &$\frac{\sum_{w \in C(w_1) \cup C(w_2)} \left( P(w|w_1) \times P(w|w_2) \right) }
{\sqrt{ \sum_{w \in C(w_1)} P(w|w_1)^2 } \times \sqrt{ \sum_{w \in C(w_2)} P(w|w_2)^2 } }$    &$\checked$         &CP\\
		                    &                &                                                    &                                                                                        &    &    \\
           ${\text{\itshape Jaccard}}^{\text{\itshape CP}}$                                 &X     &n.a.                                              &$\frac{\sum_{w \in C(w_1) \cup C(w_2)} \min (P(w|w_1),P(w|w_2))}{ \sum_{w \in C(w_1) \cap C(w_2)} \max (P(w|w_1),P(w|w_2))}$                       &$\checked$         &CP\\
		                    &               &                                                   &                                                                                        &    &    \\
           ${\text{\itshape Dice}}^{\text{\itshape CP}}$                                    &X     &n.a.                          &$\frac{2 \times \sum_{w \in C(w_1) \cup C(w_2)} \min (P(w|w_1),P(w|w_2))}{ \sum_{w \in C(w_1)} P(w|w_1) + \sum_{w \in C(w_2)} P(w|w_2)} $      &$\checked$         &CP\\ 
                            &               &                                                   &                                                &                   & \\ \hline
        \end{tabular}
        \end{minipage}
        \hspace{-0.20in}
        \end{table*}
\end{center}
\end{landscape}
\begin{landscape}
\begin{center}
\begin{table*}[htbp]
        \caption[Measures of distributional relatedness/similarity - II]{Measures of distributional relatedness/similarity (\itshape{continued}\/).}
        \label{tab:distributional measures of similarity - II}
        \hspace{-0.05in}
        \begin{minipage}{\textwidth}
        \begin{tabular}{|l c c l c c|} \hline
           {\bf Measure}    &{\bf Comp.}    &{\bf PCM}        &{\bf Formula}    &{\bf Sym.}    &{\bf Str.}\\ \hline \hline 

		                    &               &                                                   &                                                                                        &    &    \\
           ${\text{\itshape Hin}}_{\text{\itshape{rel}}}(w_1,w_2)$               &X              &n.a.                                           &$\sum_{w \in C(w)} \left\{ \begin{array}{l} \min(I(w,w_1), I(w,w_2)),\\ \qquad \quad \text{if}\; I(w,w_1) > 0\; \text{and}\; I(w,w_2) > 0 \\
						 \mid \max(I(w, w_1), I(w, w_2))\mid,\\ \qquad \quad   \text{if}\; I(w, w_1) < 0\; \text{and}\; I(w, w_2) < 0 \\
						 0,\\ \qquad \quad \text{otherwise} \end{array} \right.$                                &$\checked$         &PMI\\

		                    &               &                                                   &                                                                                        &    &    \\
           {\em Lin}              &X              &n.a.                                           &$\frac{\sum_{(r,w)\, \in\, T(w_{1})\, \cap\, T(w_{2})} (I(w_{1},r,w) + I(w_{2},r,w))}
            {{\sum_{(r,w')\, \in\, T(w_1)} I(w_1,r,w') + \sum_{(r,w'')\, \in\, T(w_2)} I(w_2,r,w'')}}$             &$\checked$         &PMI\\
                            &               &                                                    &                                                &                   & \\ 
           {\em Saif}       &X              &n.a.                                           &$\frac{2 \times \sum_{(r,w)\, \in\, T(w_{1})\, \cap\, T(w_{2})} \min(I(w_{1},r,w), I(w_{2},r,w))}
           {\sum_{(r,w') \in T(w_1)} I(w_1,r,w') + \sum_{(r,w'') \in T(w_2)} I(w_2,r,w'')}$      &$\checked$    &PMI\\ 
                            &               &                                                    &                                                &                   & \\ 
           {\em CRMs}            &X              &n.a.                                           &$\gamma \biggl[ \frac{2 \times P \times R}{P + R} \biggr]  +  (1 - \gamma) \biggl[ \beta [ P ] + (1 - \beta) [R] \biggr]$    &X        &both\footnote{The MI-based CRMs use pointwise mutual information, while the type- and token-based CRMs use conditional probability as the strength of association.} \\
                            &               &                                                    &                                                &                   & \\ \hline
        \end{tabular}
        \end{minipage}
        \hspace{-0.20in}
        \end{table*}
\end{center}
\end{landscape}

\acknowledgements
\bb{
We would like to thank Dr. Suzanne Stevenson, Dr. Gerald Penn, and Dr. Ted Pedersen for
their valuable feedback and thought-provoking discussions. This research is financially
supported by the Natural Sciences and
Engineering Research Council of Canada and the University of Toronto.}

\theendnotes

\bibliography{references}

\end{article}
\end{document}